\documentclass[11pt]{article}

\usepackage[preprint]{acl}

\usepackage{times}
\usepackage{latexsym}

\usepackage[T1]{fontenc}

\usepackage[utf8]{inputenc}

\usepackage{microtype}

\usepackage{inconsolata}

\usepackage{graphicx}

\usepackage{amsmath}
\usepackage{newtxmath}
\usepackage{algorithm}
\usepackage{algpseudocode}
\usepackage{tikz}
\usetikzlibrary{arrows.meta}
\usepackage{booktabs}
\usepackage{graphicx}
\usepackage[table]{xcolor}
\usepackage{fvextra}
\usepackage{enumitem} 
\usepackage[most]{tcolorbox}
 
\providecommand{\sd}[1]{\textsubscript{$\pm$#1}}
\usepackage[subtle]{savetrees}

\definecolor{purpleGreek}{HTML}{800080}
\definecolor{blueGreek}{HTML}{4891e7}
\definecolor{redGreek}{HTML}{d9381e}
\definecolor{greenGreek}{HTML}{088f8f}
\definecolor{yellowGreek}{HTML}{E0BC00}
\definecolor{orangeGreek}{HTML}{FFA500}
\definecolor{royalBlue}{HTML}{615CED}
\definecolor{qwenBlue}{HTML}{615CED} 

\newcommand{\mc}{{\color{blueGreek}\boldsymbol{\mu}}}
\newcommand{\qwen}{{\color{qwenBlue}\boldsymbol{\tau}}}
\newcommand{\uom}{{\color{purpleGreek}\boldsymbol{\pi}}}

\newcommand{\sufe}{{\color{orangeGreek}\boldsymbol{\zeta}}}

\definecolor{upgreen}{RGB}{46,139,87}
\definecolor{downred}{RGB}{214,69,65}

\usepackage{silence}
\tcbset{
  promptstyle/.style={
    enhanced, colback=white, colframe=black!55,
    colbacktitle=black!7, coltitle=black,
    boxrule=0.5pt, arc=1.5pt,
    left=4pt, right=4pt, top=3pt, bottom=3pt,
    toptitle=1.5pt, bottomtitle=1.5pt,
    fonttitle=\footnotesize\bfseries,
    fontupper=\footnotesize, fontlower=\footnotesize,
    before upper={\raggedright\setlength{\parskip}{2pt}\setlist{nosep,leftmargin=*}},
    before lower={\raggedright\setlength{\parskip}{2pt}\setlist{nosep,leftmargin=*}},
    sidebyside gap=12pt, lefthand ratio=0.5, sidebyside align=top seam,
    segmentation style={black!25},
  },
  promptraster/.style={
    raster columns=2, raster equal height=rows,
    raster column skip=6pt, raster row skip=6pt,
  },
}
\newtcolorbox{promptpanel}[2][]{promptstyle, title={#2}, #1}
\newcommand{\ph}[1]{\texttt{\{#1\}}}  
\newcommand{\phead}[1]{\par\smallskip\noindent\textit{#1}\par}  
\fvset{fontsize=\scriptsize,breaklines}
\usepackage{arydshln}
\title{Which Agent's Skill Should Change? \\Group-Relative Credit Assignment for Evolving Multi-Agent Teams}

\author{
  \parbox{\textwidth}{\centering
    \textbf{Yanan Ma}$^{\mc}$%
    \thanks{Corresponding authors' email addresses:
    \texttt{\{ma.yanan;banerjee.imon\}@mayo.com}. Our code is available at \texttt{https://github.com/yananma1029/GRSE-Skills}.}
    \textbf{Yang Wang}$\hspace{.1em}^{\uom}$,
    \textbf{David Le}$\hspace{.1em}^{\mc}$,
    \textbf{Chenhan Yuan}$\hspace{.1em}^{\qwen}$,
    \textbf{Yizheng Sun}$\hspace{.1em}^{\uom}$, \\
    \textbf{Guangwei Zhang}$\hspace{.1em}^{\mc}$,
    \textbf{Chenghao Xiao}$\hspace{.1em}^{\sufe}$,
    \textbf{Chenghua Lin}$\hspace{.1em}^{\uom}$,
    \textbf{Imon Banerjee}$\hspace{.1em}^{\mc}$
  \\ \vspace{0.4cm}
    \textnormal{$\hspace{.1em}^{\mc}$Mayo Clinic},
    \textnormal{$\hspace{.1em}^{\uom}$University of Manchester},\\
    \textnormal{$\hspace{.1em}^{\qwen}$Alibaba Token Hub, Alibaba Group},
    \textnormal{$\hspace{.1em}^{\sufe}$Shanghai University of Finance and Economics}
  }
}

\begin{document}
\maketitle
\begin{abstract}
Cooperative language-model agents can improve without weight updates by revising reusable skills, but only if a shared team reward is traced to the responsible agent and skill. Existing methods credit the skills present in each trajectory, yet presence is not responsibility: a harmful skill active in successful runs still earns positive credit. We present \textbf{GRSE}, \textbf{G}roup-\textbf{R}elative \textbf{S}kill \textbf{E}volution, which assigns behavioral credit by comparison: inspired by GRPO, it executes each question several times under the same skill libraries and contrasts the executions to credit agent--skill pairs, with structured diagnosis of failures. On HotpotQA and LoCoMo with two open-weight actors, GRSE outperforms MASkills, the closest prior method, on HotpotQA and every LoCoMo category except open-domain, with the largest gains on multi-hop questions ($+6.7$ and $+3.2$ F1). Its gains rest on better-grounded retrieval rather than answers given without evidence, and its learned skills transfer between the multi-hop tasks. GRSE is robust to corrupted libraries: starting from planted harmful skills, it credits every planted skill with the correct sign and recovers the score of a clean library ($+6.3$ points), whereas MASkills misattributes two and falls below its starting point.
\end{abstract}
\section{Introduction}
\label{sec:introduction}

\begin{figure*}[t]
\centering
\includegraphics[width=0.9\textwidth]{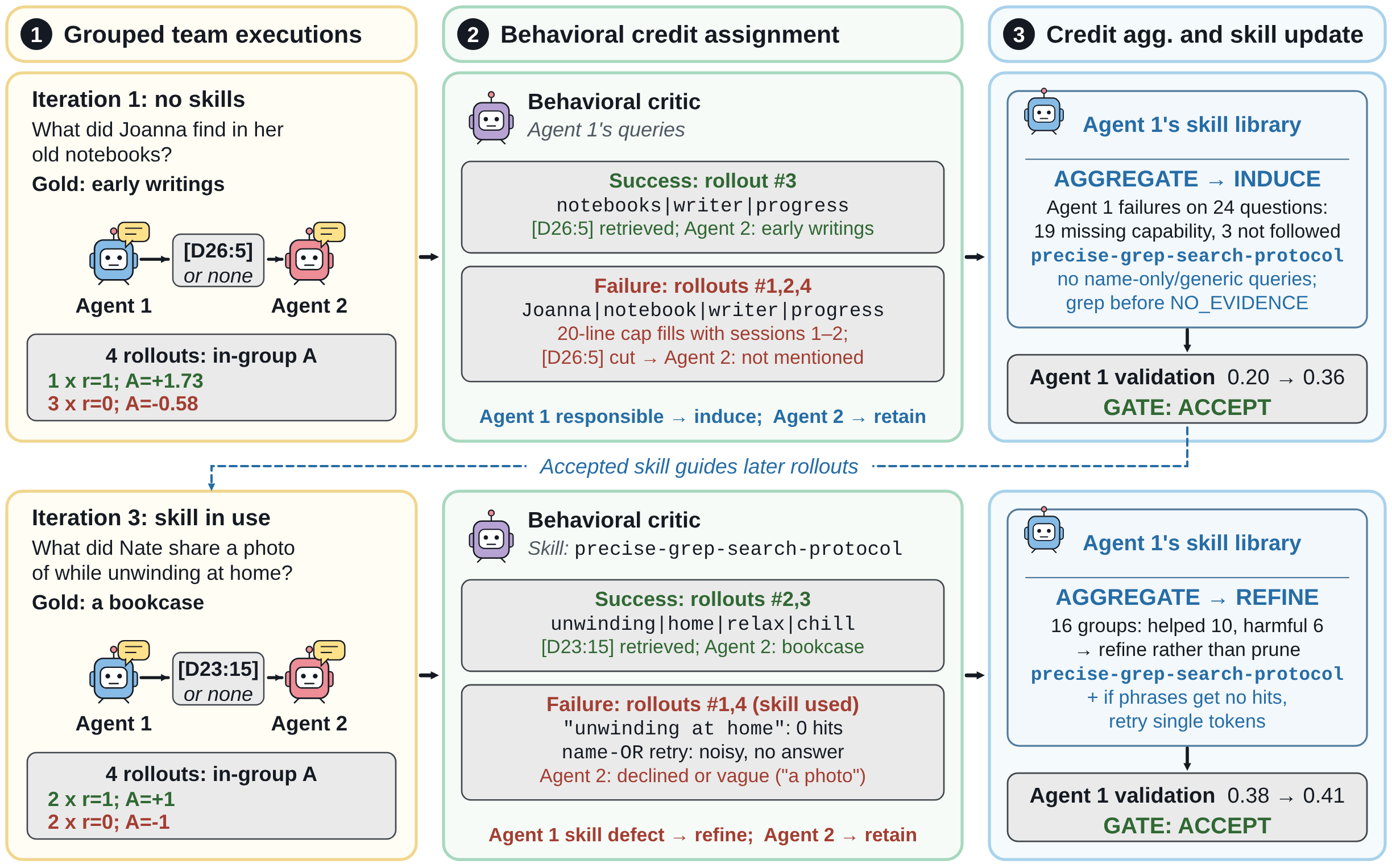}

\caption{Each iteration (1) samples a group of team executions per question (\S\ref{sec:method-groups}), (2) credits agent--skill pairs by contrasting these executions and diagnoses failures in a structured form (\S\ref{sec:method-credit}), and (3) aggregates the credit across questions into skill edits, each kept only if it does not lower the validation score (\S\ref{sec:method-organization}--\ref{sec:method-evolution}). The example is from one training run on LoCoMo single-hop questions, in which Agent~1 retrieves evidence and Agent~2 answers. In iteration~1, a failure that surfaces in Agent~2's answer is traced to Agent~1's retrieval, and a new retrieval skill is induced. In iteration~3, this skill is credited as harmful in the shown group but helpful across the iteration's 16 groups, so it is refined rather than pruned. In Agent~1's search queries, \texttt{|} denotes OR.}
\label{fig:framework}
\end{figure*}

Language-model agents increasingly work in teams: one agent collects evidence, another reasons over it, and the team is judged by a single outcome~\citep{cemri2026multi,ghareeb2026multi,tran2025multi,dang2026multi,zhang2026g,xia2026mmedagent,hong2024metagpt,wu2023autogen,li2023camel}. Improving such a team requires answering a credit assignment question: which agent's behavior, and which step of it, produced a success or a failure? The question is classical in multi-agent reinforcement learning, where a shared reward is decomposed with difference rewards or counterfactual baselines~\citep{wolpert2001optimal,foerster2018counterfactual,rashid2020monotonic}, and it has recently been posed for language-model teams as automated failure attribution~\citep{zhang2025agent}.  A wrong answer may stem from evidence collection, an incomplete handoff, or downstream reasoning~\citep{wang2026flat,zhang2025agent}, and the agent whose output is wrong is often not the agent whose behavior needs to change. For example, reasoner answers wrongly when the retriever misses the evidence.

We study this problem in a setting where model parameters stay fixed and each agent's behavior is specified by natural-language skills, reusable procedures that the agent reads and follows~\citep{zhao2024expel,wang2023voyager,alzubi2026evoskill,yao2026maskills}. Skills make behavior editable, so credit becomes actionable: once responsibility is located, the responsible procedure can be revised. Existing experience- and skill-learning methods, however, assign credit coarsely, which often distill lessons from whole trajectories, and mainly focus on single-agent setting~\citep{shinn2023reflexion,xu2026mem,ouyang2026reasoningbank}. Meanwhile MASkill, the recent two-agent SOTA, credits each skill from single trajectory in isolation~\citep{yao2026maskills}. However, without a contrast, a skill that is present in every execution also receives credit for the successes it did not cause: in our robustness study, MASkills raises the utility of a planted harmful skill over three iterations, whereas our method labels it harmful and prunes it (Appendix~\ref{sec:robustness-analysis}). Comparison between executions already guide experience distillation, skill refinement, and prompt optimization~\citep{cai2025training, chen2026skillcat, li2026skillcome,zehle2026cantante}, but they have not been used to localize a team's outcome to individual agents' behaviors.

Behavioral credit assignment in agent teams poses two challenges. First, attribution must look past where an error surfaces. In Figure~\ref{fig:framework}, Agent~2 answers ``not mentioned'' in three of four executions of the same question, yet Agent~2 behaves correctly given its input. The decisive difference lies in Agent~1's search query, which retrieves the answer only in the successful execution. Second, an attribution must be specific and checkable before it can drive a revision. A defective procedure, a missing capability, and a tool failure call for different responses, and an explanation that cites no execution step cannot be verified. Across questions, attributions must also be aggregated without counting repeated executions of one question as independent evidence.

We introduce \textbf{GRSE}, \textbf{G}roup-\textbf{R}elative \textbf{S}kill \textbf{E}volution, a training-free framework that evolves cooperative agents' skills through group relative behavioral credit assignment. Following GRPO~\citep{shao2024deepseekmath}, GRSE samples multiple complete team rollouts of the same question under fixed skill libraries and compute relative within-group advantages (\S\ref{sec:method-groups}). A group-relative critic compares agent behavior and handoffs across these rollouts to assign feedback to individual agent--skill pairs. A structured diagnosis grounds each attribution in a failure type, cited failed and successful steps, and an edit action, and reference checks reject attributions that the executions do not support (\S\ref{sec:method-credit}). The accepted attributions are aggregated across questions (\S\ref{sec:method-organization}) and applied as skill updates through the editing operators and validation gate of MASkills (\S\ref{sec:method-evolution}). Figure~\ref{fig:framework} shows both levels of attribution within a single run: \emph{across agents}, responsibility shifts from the answering agent to the retrieving agent; \emph{within an agent}, a previously learned skill is later credited as harmful and diagnosed as a skill defect. Our contributions are threefold:


\begin{itemize}
    \item We formulate skill evolution for cooperative agents as \emph{behavioral credit assignment}: with model parameters fixed and skills as editable units, a shared team outcome must be attributed to the agent and skills responsible for it. We address this by jointly comparing same-question team executions, which produces agent--skill feedback while preserving the handoffs needed to interpret each agent's behavior and the proposed revision.
    \item We make these attributions verifiable and actionable: a structured failure analysis links each attribution to a failure type, checked step references, and an edit action, and cross-question aggregation weighs attributions by the number of distinct questions that support them.
    \item We evaluate GRSE on HotpotQA and LoCoMo with two open-weight actors. GRSE outperforms MASkills on HotpotQA and every LoCoMo category except open-domain. Its credit identifies and removes harmful skills that MASkills retains. Answer-transition and learned-skill analyses show that GRSE's gains come from fewer broken answers and better-grounded retrieval.
\end{itemize}
\section{Related Work}
\label{sec:related-work}

\subsection{Reflection and Experience Memory}
Experience-based adaptation stores information from past interactions to guide future decisions. Reflexion~\citep{shinn2023reflexion} records verbal reflections for subsequent attempts. ExpeL~\citep{zhao2024expel} extracts reusable insights from successes and failures, including comparisons on the same task. A-MEM~\citep{xu2026mem} constructs linked notes and updates their contextual attributes as new memories arrive. ReasoningBank~\citep{ouyang2026reasoningbank} distills reasoning strategies, and its memory-aware test-time scaling uses multiple executions to enrich memory curation. These methods actively transform experience rather than simply retaining raw traces. Memory and skills also interact: MemSkill~\citep{zhang2026memskill} learns to select memory-management skills and revises its skill bank, while MSCE~\citep{tang2026memory} promotes evidence-supported procedural policies into skills and maintains them using feedback. These methods decide \emph{what} experience to keep and how to organize it, whereas GRSE instead decides \emph{where} execution evidence should act, assigning it to editable skills owned by individual agents.

\subsection{Reusable Skill Discovery and Evolution}
Skill libraries turn experience into procedures that can be invoked and revised. Voyager~\citep{wang2023voyager} maintains executable code skills, and EvoSkill~\citep{alzubi2026evoskill} uses an executor, proposer, and skill builder to discover and refine skills from task feedback. SkillRL~\citep{xia2026skillrl} distills failed trajectories into lessons, organizes skills into general and task-specific categories, and evolves the library alongside reinforcement learning of the agent policy. These methods learn skills for a single executing agent, so they never have to decide which agent's skill to revise. MASkills~\citep{yao2026maskills} extends skill learning to cooperative teams through skill-conditioned credit, hierarchical aggregation, and skill editing, but it credits each trajectory on its own. We retain its skill representation and editing operators and change how credit is formed: GRSE jointly compares same-question team executions, and structured diagnosis links each failure to an agent--skill target, cited execution steps, and a checked edit action.

\subsection{Group-Relative and Contrastive Feedback}
GRPO~\citep{shao2024deepseekmath} uses within-question rewards to optimize model parameters. Training-Free GRPO~\citep{cai2025training} instead distills semantic feedback from rollout groups into experiential context. Several methods already apply comparisons to skills: SkillCAT~\citep{chen2026skillcat} contrasts same-task success/failure pairs, evaluates patches through replay, and organizes skills for routing. SkillCome~\citep{li2026skillcome} combines within-group contrast with memories of failure and contrast patterns. CANTANTE~\citep{zehle2026cantante} contrasts joint prompt configurations on the same query to provide per-agent optimization signals. In these methods, the contrast either compares outcomes or varies the configuration being optimized. GRSE holds the skills fixed and contrasts the team's behavior across executions, so that an outcome difference can be traced to an agent, a skill, and a cited step.
\section{Method}
\label{sec:method}


GRSE attributes a team's outcome to the behaviors of individual agents and turns these attributions into revisions of the responsible behavior. Each agent's behavior is specified by a skill library, while language-model parameters and agent roles stay fixed. Every iteration has three main steps (Figure~\ref{fig:framework}): grouped rollouts provide contrasting executions of the same question (\S\ref{sec:method-groups}); a group-relative critic assigns behavioral credit to agent--skill pairs and grounds it in structured failure analysis (\S\ref{sec:method-credit}); and the attributions are aggregated across questions (\S\ref{sec:method-organization}) and applied as validated skill updates (\S\ref{sec:method-evolution}). Algorithm~\ref{alg:skill-evolution} summarizes the update loop.  

\subsection{Grouped Rollouts and Relative Feedback}
\label{sec:method-groups}

For $N$ cooperating agents, let $\mathcal{K}^{t}=(K_1^{t},\ldots,K_N^{t})$ be their skill libraries at iteration $t$. Each library contains named instruction artifacts following MASkills~\citep{yao2026maskills}. A complete team trajectory $\tau$ records agent actions, messages, tool interactions, and the final answer. Its skill trace $\xi$ identifies candidate agent--skill attribution targets. Trace membership records exposure, while the behavioral evidence supports the critic assessment.

For each question $q$ in batch $\mathcal{B}_t$, we sample $G$ complete team executions from a fresh context under the same libraries and form
\begin{equation}
    \mathcal{G}_q^{t}
    =\{(\tau_{qg},r_{qg},A_{qg},\xi_{qg})\}_{g=1}^{G},
    \label{eq:rollout-group}
\end{equation}
where $r_{qg}=R(q,\tau_{qg})\in[0,1]$. We set $G=4$ consistently in our experiments. Following the normalization in GRPO~\citep{shao2024deepseekmath}, let $\bar r_q=G^{-1}\sum_g r_{qg}$ and $\sigma_q^2=G^{-1}\sum_g(r_{qg}-\bar r_q)^2$. The relative advantage is
\begin{equation}
    A_{qg}=\begin{cases}
        \dfrac{r_{qg}-\bar r_q}{\sigma_q+\epsilon}, & \sigma_q>\epsilon,\\[3pt]
        0, & \sigma_q\leq\epsilon.
    \end{cases}
    \label{eq:group-advantage}
\end{equation}

Because all executions in a group share the question and the libraries, the differences between high- and low-advantage executions point to the behaviors that decided the outcome, much like a counterfactual baseline in multi-agent reinforcement learning~\citep{foerster2018counterfactual}. The critic receives both absolute rewards $r_{qg}$ and relative advantages $A_{qg}$ to compare trajectories and draws on execution evidence to assess skill credit (\S\ref{sec:method-credit}). Groups with equal-reward carry no contrasts but still show which behaviors consistently fail or succeed, so they are kept by default. We further tested groups with differing rewards following DAPO~\citep{yu2026dapo} in our ablation studies (\S\ref{sec:experiment-ablation}).

\subsection{Group-Relative Behavioral Critic Assignment}
\label{sec:method-credit}

The critic reads all $G$ executions of a question together and compares them agent by agent. It can therefore attribute an outcome difference to the agent whose behavior differs between high- and low-advantage executions, rather than to the agent whose output is wrong:

\begin{equation}
    \operatorname{Critic}(q,\mathcal{G}_q^{t},\mathcal{K}^{t})
    = (\mathcal{C}_q,B_q,\mathcal{D}_q).
    \label{eq:group-critic}
\end{equation}

Here $\mathcal{C}_q=\{c_{qik}\}_{i,k}$ contains at most one skill credit per relevant agent $i$ and skill $k$. $B_q$ records agent-level feedback not tied to an existing skill and evidence for skill induction. $\mathcal{D}_q$ contains structured failure diagnoses of existing skills with checked execution references and edit actions. In Figure~\ref{fig:framework}, iteration~1 yields agent-level credit for Agent~1, since no skill exists yet, and iteration~3 yields a skill credit for Agent~1's learned retrieval skill.

\paragraph{Skill credit.} For each credit $c_{qik}=(v,e,\Delta,\mathcal{R},\mathcal{F},u)_{qik}$, $v$ labels the skill as helped, redundant, harmful, or neutral; $e$ gives supporting evidence; and $\Delta$ suggests a refinement. $\mathcal{R}$ and $\mathcal{F}$ identify redundant and conflicting skills, respectively. The critic's utility score $u\in[-1,1]$ suggests skill usefulness and informs cross-question aggregation and subsequent pruning decisions, while redundancy links guide consolidation. With empty libraries, $B_q$ or $\mathcal{D}_q$ supplies evidence for induction.

\paragraph{Structured failure analysis} A credit states \emph{who} is responsible, and a failure analysis states \emph{why} and \emph{where}, in a form that can be checked. We represent it as $d=(i,k,z,L^{-},L^{+},e,\Delta,a)$: agent, skill identifier, failure type, failed-step references, optional successful-step references, explanation, proposed correction, and action, respectively. The critic selects $z$ from eight predefined failure types: \textit{skill not followed}, \textit{skill defect}, \textit{missing capability}, \textit{handoff error}, \textit{evidence missing}, \textit{answer format}, \textit{tool failure}, and \textit{uncertain attribution}. These types distinguish procedural defects from execution errors and failures that do not justify a skill edit. The action $a$ is \textit{refinement}, \textit{induction}, or \textit{no edit}. Induction uses $k=\varnothing$ to indicate a missing capability. 

\paragraph{Attribution checks.} The critic interface checks that step references exist, match the named agent, and cite non-full-reward executions for $L^{-}$ and full-reward executions for $L^{+}$. Refinement additionally requires an owned skill recorded in the cited failed rollout and a matching credit. Tool failures and uncertain or unknown responsibility permit only no edit. Accepted diagnoses provide the failure-related edit directions in $\mathcal{C}_q$ and the induction evidence in $B_q$. A correction without a successful comparison is marked as a hypothesis. These checks enforce reference and action consistency. The explanation remains an observational assessment. In Figure~\ref{fig:framework}, iteration~3, the analysis names Agent~1 and its learned skill, classifies the failure as a skill defect, cites the failed rollouts \#1 and \#4 and the successful rollouts \#2 and \#3, and proposes a refinement.

\subsection{Cross-Question Credit Aggregation}
\label{sec:method-organization}

A single group shows how one question went and a revision should rest on attributions that recur across questions. We therefore aggregate credit across questions into edit direction for existing skills and evidence for new skills. For existing skills, feedback for $s_{ik}^{t}\in K_i^t$ is synthesized into an edit direction:
\begin{equation}
    \Gamma_{ik}=\operatorname{Aggregate}
    \bigl(s_{ik}^{t},\{c_{qik}:q\in\mathcal Q_{ik}\}\bigr),
    \label{eq:skill-aggregation}
\end{equation}
where $\mathcal Q_{ik}$ contains questions with usable credit for that pair. We reuse the MASkills~\citep{yao2026maskills} aggregation operator, which returns a summary, suggested edit, utility score, and verdict. Let $\Gamma_i$ collect these outputs, augmented with category utility summaries when category is enabled.

For skill induction, the organizer groups accepted diagnoses across questions by agents, failure type, skill target, and edit action. Each pattern $p=(i,z,k,a)$ is assigned support
\begin{equation}
    w(p)=\left|\left\{q:\exists d\in\mathcal{D}_q,
    \operatorname{key}_q(d)=p\right\}\right|,
    \label{eq:pattern-support}
\end{equation}
where $\operatorname{key}_q$ extracts the grouping fields. For each agent, induction patterns are ordered by $w(p)$ and retained with example diagnoses as evidence $E_i$. Thus repeated rollouts do not count as additional supporting questions. The aggregated feedback $\Gamma_i$ guides edits to existing skills, while $E_i$ provides evidence for inducing new skills.

\subsection{Credit-Guided Skill Evolution}
\label{sec:method-evolution}

Given the aggregated edit directions $\Gamma_i$ and induction evidence $E_i$, we evolve each agent's skill library. With operator schedule $\Omega$ and library limit $L$, candidate libraries are
\begin{equation}
    \widehat K_i=\operatorname{Evolve}
    \bigl(K_i^t,\Gamma_i,E_i;\Omega(t+1),L\bigr).
    \label{eq:skill-evolution}
\end{equation}

The operators refine credited skills along $\Gamma_i$, prune skills with low utility, consolidate redundant ones, and induce new skills from $E_i$. Each candidate is validated against the same pre-update team, so that an update whose attribution was wrong is not kept. For fixed validation set $\mathcal V$ and $H$ executions per question, define
\begin{equation}
    S(\mathcal K)=\frac{1}{|\mathcal V|H}
    \sum_{q\in\mathcal V}\sum_{h=1}^{H}R(q,\tau_{qh}^{\mathcal K}).
    \label{eq:validation-score}
\end{equation}
Let $b_t=S(\mathcal K^t)$ and $\widehat s_i=S(\mathcal K^t_{i\leftarrow\widehat K_i})$, where only agent $i$'s library is replaced. Tolerance $\delta$ gives the update
\begin{equation}
    K_i^{t+1}=\begin{cases}
        \widehat K_i, & \widehat s_i\geq b_t-\delta,\\[2pt]
        K_i^{t}, & \text{otherwise}.
    \end{cases}
    \label{eq:validation-gate}
\end{equation}
Accepted candidates are assembled after all agents are tested, and acceptance remains per agent. 

\begin{algorithm}[ht]
\caption{Group-Relative Skill Evolution}
\label{alg:skill-evolution}
\small
\begin{algorithmic}[1]
\State \textbf{Input:} training tasks $\mathcal D$, validation set $\mathcal V$,
    initial libraries $\mathcal K^0$, iterations $T$, group size $G$,
    tolerance $\delta$, schedule $\Omega$, library limit $L$.
\State \textbf{Output:} evolved libraries $\mathcal K^T$.
\For{$t=0,\ldots,T-1$}
    \State Sample $\mathcal B_t\subseteq\mathcal D$.
    \ForAll{$q\in\mathcal B_t$}
        \State Collect $\mathcal G_q^t$ with fixed $\mathcal K^t$ (Eqs.~\ref{eq:rollout-group}--\ref{eq:group-advantage}).
        \State Obtain usable $f_q=(\mathcal C_q,B_q,\mathcal D_q)$ via Eq.~\eqref{eq:group-critic}.
    \EndFor
    \ForAll{agents $i=1,\ldots,N$}
        \State Aggregate skill feedback $\Gamma_i$ via Eq.~\eqref{eq:skill-aggregation}.
        \State Build $E_i$ from $B_q$ and low-reward groups, or structured patterns (Eq.~\eqref{eq:pattern-support}).
        \State Append structured category utility summaries to $\Gamma_i$ when enabled.
        \State Form candidate $\widehat K_i$ via Eq.~\eqref{eq:skill-evolution}.
    \EndFor
    \State Compute $b_t=S(\mathcal K^t)$ via Eq.~\eqref{eq:validation-score}.
    \ForAll{agents $i=1,\ldots,N$}
        \State If unchanged, retain $K_i^t$; otherwise test $\widehat K_i$ against $\mathcal K^t$ and apply Eq.~\eqref{eq:validation-gate}.
    \EndFor
    \State Assemble, evaluate, and save $\mathcal K^{t+1}$.
\EndFor
\State \Return $\mathcal K^T$.
\end{algorithmic}
\end{algorithm}

\section{Experiment}
\label{sec:experiment}
\subsection{Experimental Setup}
\label{sec:experiment-setup}

We follow the setting of MASkills~\citep{yao2026maskills}, the closest prior SOTA for multi-agent skill evolution. Each method consists of two agents with task-specific roles that communicate sequentially, and the second agent produces the final answer. The agents are executed by one of two open-weight actors, Qwen3.8-27B~\citep{qwen38} and Gemma-4-31B-it~\citep{{gemmateam2026gemma4}}, each conditioned on its own skill library. We use these two recent released open-weight model of similar size from different families to test whether credit assignment depends on the actor. A fixed Qwen3.8-27B model serves as critic and skill optimizer for all methods. Model parameters and agent roles stay fixed. The agents change only through their skills, which are revised according to the assigned credit.

All skill-evolution methods start from empty libraries and run for five iterations. GRSE samples four team trajectories per question, whereas MASkills samples one. Final libraries are evaluated with temperature-zero decoding. Detailed model, data and optimization settings appear in Appendix~\ref{sec:app:implementation}.

\subsection{Benchmarks}
\label{sec:experiment-datasest}
\label{sec:experiment-baseline}

\paragraph{Datasets and metrics.} We evaluate on HotpotQA~\citep{yang2018hotpotqa} for multi-hop question answering and LoCoMo~\citep{maharana2024evaluating} for long-memory conversational reasoning. We use the distractor--hard subset and report the answer-token F1 for HotpotQA. LoCoMo covers multi-hop, temporal, open-domain, and single-hop questions. We report the category-specific F1 and smoothed BLEU. 

\paragraph{Baselines.} We group baselines by whether they use the agent team. \emph{Single-model baselines} answer with one model and no agents or skills. On HotpotQA, we run four prompting methods with the same actor, questions, passages, answer-format instruction, and metric as the \emph{agent teams}, including IO (direct answer), Chain-of-Thought(CoT;~\citep{wei2022chain}), CoT with self-consistency (CoT-SC;~\citep{wang2022self}; majority vote over five samples), and Self-Refine~\citep{madaan2023self} (at most three feedback-and-revision rounds). On LoCoMo, the full-context baseline follows the benchmark protocol~\citep{maharana2024evaluating} with the actor model reads the entire conversation and answers directly. \emph{Two-agent baselines} share the team, roles, and actor with GRSE. The empty-library team is this team before any skill is learned (iteration~0). For HotpotQA, all teams follow the agent prompts and short-answer instructions from the MASkills implementation, with questions and distractor passages supplied by the dataset. For LoCoMo, agent teams use category-specific task prompts. All the prompts are detailed in Appendix~\ref{sec:app:template}. We report each \emph{two-agent} method as the mean over three training runs, because the learned libraries vary with the sampled rollouts and the critic's outputs, whereas \emph{single-model} baselines have no training stage and are evaluated once.

\paragraph{GRSE variants.} We report GRSE and \emph{GRSE w/o structure analysis}, which retains group-relative feedback but omits the prescribed failure-diagnosis schema and its checks, in main analysis. We report \emph{GRSE w/o structure} and the other three variants in ablation studies: \emph{GRSE w/ label} retains structured diagnosis and additionally uses dataset labels to condition the critic and organize update evidence; \emph{GRSE with $G{=}1$} samples one trajectory per question; and \emph{DAPO filter} keeps four rollouts but, following the dynamic sampling of DAPO~\citep{yu2026dapo}, sends only groups whose rewards differ to the critic. 

\section{Results and Discussion}
\subsection{Main Results}
\begin{table}[ht]
\centering
\small
\setlength{\tabcolsep}{5pt}
\begin{tabular}{lcc}
\toprule
Actor Model & Qwen3.8 & Gemma4 \\
\midrule
\multicolumn{3}{l}{\textit{Single model, no agents}} \\
\quad IO & 81.6 & 81.3\\
\quad CoT & 80.7 & 82.5\\
\quad CoT-SC & 82.6 & \textbf{83.3}\\
\quad Self-Refine & \textbf{84.1} & 81.1\\
\midrule
\multicolumn{3}{l}{\textit{Two agents, 5 iterations}} \\
\quad Empty library (iter.~0) & 80.6 & 81.6 \\
\quad MASkills & 80.6\sd{0.6} & 82.3\sd{0.2} \\
\rowcolor{gray!12}\quad GRSE (ours) & \underline{83.7}\sd{0.5} & 82.8\sd{0.4} \\
\rowcolor{gray!12}\quad GRSE w/o structure & 83.1\sd{0.1} & \textbf{\underline{83.3}}\sd{0.4} \\
\bottomrule
\end{tabular}
\caption{HotpotQA test F1 ($\times$100) with a greedy actor on 500 test questions. Columns give the actor model (Qwen3.8-27B, Gemma-4-31B-it). Qwen3.8-27B is critic and optimizer throughout. Empty library is this system before any skill is learned (iteration~0). Mean over three runs, with SD as a subscript. CoT-SC uses 5 samples, Self-Refine at most 3 feedback rounds. Best per column in \textbf{bold}. Best two-agent setting is \underline{underlined}.}
\label{tab:main-hotpotqa}
\end{table}

\begin{table*}[t]
\centering
\small
\setlength{\tabcolsep}{4.5pt}
\begin{tabular}{@{}lcccccccc@{}}
\toprule
 Task ($N$) &  \multicolumn{2}{c}{Multi-hop (156)} & \multicolumn{2}{c}{Temporal (179)} & \multicolumn{2}{c}{Open-domain (52)} & \multicolumn{2}{c}{Single-hop (453)} \\
\cmidrule(lr){2-3} \cmidrule(lr){4-5} \cmidrule(lr){6-7} \cmidrule(l){8-9}
Method  & F1 & BLEU & F1 & BLEU & F1 & BLEU & F1 & BLEU \\
\midrule
\multicolumn{9}{@{}l}{\textit{Actor: Qwen3.8-27B}} \\
Full-context            & 49.9 & 40.0 & 48.3 & 25.9 & \textbf{36.4} & \textbf{31.6} & \textbf{70.8} & 57.6 \\
\cdashline{1-9}\noalign{\vskip 2pt}
Empty library (iter.~0) & 49.7 & 42.8 & 70.6 & 48.5 & 24.7 & 16.4 & 59.7 & 51.0 \\
MASkills & 46.7\sd{1.4} & 39.4\sd{0.8} & 72.8\sd{1.3} & 51.8\sd{2.1} & \underline{31.8}\sd{4.4} & \underline{22.8}\sd{5.1} & 59.0\sd{0.6} & 49.5\sd{1.2} \\
\rowcolor{gray!12}
GRSE (ours) & 53.4\sd{0.8} & 44.7\sd{0.7} & \textbf{\underline{73.4}}\sd{2.1} & \textbf{\underline{52.9}}\sd{2.2} & 24.7\sd{1.1} & 15.9\sd{1.3} & 59.1\sd{4.5} & 50.0\sd{4.1} \\
\rowcolor{gray!12}
GRSE w/o structure & \textbf{\underline{54.9}}\sd{0.9} & \textbf{\underline{46.6}}\sd{0.7} & 71.1\sd{1.9} & 49.5\sd{2.3} & 29.1\sd{1.6} & 20.1\sd{1.5} & \underline{67.7}\sd{3.5} & \textbf{\underline{58.0}}\sd{3.8} \\
\midrule
\multicolumn{9}{@{}l}{\textit{Actor: Gemma~4-31B}} \\
Full-context            & \textbf{48.5} & \textbf{37.8} & 55.7 & 30.0 & 18.6 & 12.9 & \textbf{71.7} & \textbf{57.6} \\
\cdashline{1-9}\noalign{\vskip 2pt}
Empty library (iter.~0) & 36.5 & 29.7 & 65.6 & 45.1 & 19.2 & 10.7 & 45.3 & 37.0 \\
MASkills & 37.9\sd{4.5} & 30.8\sd{3.2} & 61.3\sd{2.4} & 41.2\sd{1.6} & \textbf{\underline{32.2}}\sd{5.6} & \textbf{\underline{25.2}}\sd{4.1} & 46.6\sd{2.0} & 37.6\sd{2.0} \\
\rowcolor{gray!12}
GRSE (ours) & \underline{41.1}\sd{3.8} & \underline{33.6}\sd{3.0} & \textbf{\underline{67.4}}\sd{1.4} & \textbf{\underline{45.9}}\sd{1.5} & 30.4\sd{6.4} & 21.7\sd{7.6} & 54.0\sd{6.5} & 45.2\sd{3.8} \\
\rowcolor{gray!12}
GRSE w/o structure & 38.2\sd{3.9} & 31.2\sd{3.3} & 65.9\sd{1.9} &{45.7}\sd{3.2} & 30.4\sd{5.4} & 21.7\sd{6.3} & \underline{57.1}\sd{0.5} & \underline{46.3}\sd{0.9} \\
\bottomrule
\end{tabular}
\caption{LoCoMo test results per question category: token F1 and smoothed BLEU ($\times$100) with a greedy actor.  Full-context is a single model that receives the entire conversation, following the LoCoMo protocol (no agents or skills). The remaining rows use the two-agent system trained for 5 iterations. Empty library is this system before any skill is learned (iteration~0). Mean over three runs, with SD as a subscript. Qwen3.8-27B serves as critic and optimizer for both skill methods. Best per column in \textbf{bold}; best two-agent result \underline{underlined}.}
\label{tab:main-locomo}
\end{table*}

Overall, GRSE is the strongest two-agent method on HotpotQA with both actors, outputs MASkills on every LoCoMo category except open-domain, and varies less across runs than MASkill (Tables~\ref{tab:main-hotpotqa} and~\ref{tab:main-locomo}).

\paragraph{HotpotQA.} GRSE improves over MASkills with both actors: 83.7 vs.\ 80.6 F1 with Qwen3.8 ($+3.1$) and 82.8 vs.\ 82.3 with Gemma4 ($+0.5$). With Qwen3.8, it comes within 0.4 F1 of the strongest single-model baseline, Self-Refine (84.1). With Gemma4, all methods lie within 2.2 F1 of one another (81.1--83.3), and GRSE w/o structure is the strongest two-agent setting (83.1). 

\paragraph{LoCoMo.} GRSE improves over MASkills on every LoCoMo category except open-domain with both actors. On multi-hop, it gains $+6.7$ F1 and $+5.3$ BLEU with Qwen3.8 and $+3.2$ F1 and $+2.8$ BLEU with Gemma4. On temporal questions, it achieves the highest scores with both actors ($73.4$ F1 with Qwen3.8 and $67.4$ F1 With Gemma4). On single-hop questions, GRSE w/o structure achieves the best two-agent results ($67.4$ F1 with Qwen3.8 and $57.1$ with Gemma4). Open-domain, the smallest category (52 questions), favor MASkills with both actors. Relative to the empty library, the skills learned by GRSE help where those of MASkills can hurt. With Gemma4, both GRSE variants improve over the empty library in every category (GRSE: $+1.8$ to $+11.2$ F1), whereas MASkills falls below it on temporal questions ($-4.3$ F1). With Qwen3.8, MASkills likewise falls below the empty library on multi-hop questions ($-3.0$ F1), where GRSE gains $+3.7$. On multi-hop questions, GRSE also varies less across runs than MASkills (F1 SD 0.8 vs.\ 1.4 with Qwen3.8 and 3.8 vs.\ 4.5 with Gemma4).

The full-context baseline is an upper-bound-style reference rather than a competitor: it gives a single model the entire conversation, whereas the two-agent systems must first retrieve evidence and answer from that alone. Full-context retains the advantage on single-hop questions with both actors, multi-hop scores with Gemma4, and Open-domain scores with Qwen3.8. Whereas seeing only retrieved evidence, GRSE surpasses it on temporal questions with both actors.

\subsection{Analysis of Learned Skills}
\label{sec:experiment:learned_skills}
\begin{table}[t]
\centering
\footnotesize
\setlength{\tabcolsep}{2.5pt}
\resizebox{\columnwidth}{!}{%
\begin{tabular}{@{}lrrrrccc@{}}
\toprule
 & & \multicolumn{3}{c}{Contribution to $\Delta$} & \multicolumn{2}{c}{\% of questions} & \\
\cmidrule(lr){3-5} \cmidrule(lr){6-7}
Method & \multicolumn{1}{c}{$\Delta$} & \multicolumn{1}{c}{D${\to}$C} & \multicolumn{1}{c}{W${\to}$C} & \multicolumn{1}{c}{C${\to}$W} & Decl. & No ev. & \#D${\to}$C \\
\midrule
\multicolumn{8}{@{}l}{\textit{HotpotQA, Qwen3.8}} \\
MASkills & $+0.7$ & $0.0$ & $+2.3$ & $-1.8$ & $0{\to}0$ & -- & -- \\
\rowcolor{gray!12}
GRSE     & $+2.3$ & $0.0$ & $+2.6$ & $-0.6$ & $0{\to}0$ & -- & -- \\
\cdashline{1-8}\noalign{\vskip 2pt}
\multicolumn{8}{@{}l}{\textit{HotpotQA, Gemma4}} \\
MASkills & $+0.7$ & $0.0$ & $+1.2$ & $-0.8$ & $1{\to}1$ & -- & -- \\
\rowcolor{gray!12}
GRSE     & $+1.0$ & $0.0$ & $+1.2$ & $-0.6$ & $1{\to}2$ & -- & -- \\
\midrule
\multicolumn{8}{@{}l}{\textit{LoCoMo, Qwen3.8}} \\
MASkills & $-0.5$ & $+1.8$ & $+3.3$ & $-4.8$ & $5{\to}4$ & $6{\to}5$ & 18 (0) \\
\rowcolor{gray!12}
GRSE     & $+5.9$ & $+2.4$ & $+6.3$ & $-2.4$ & $5{\to}2$ & $6{\to}5$ & 22 (0) \\
\cdashline{1-8}\noalign{\vskip 2pt}
\multicolumn{8}{@{}l}{\textit{LoCoMo, Gemma4}} \\
MASkills & $-0.3$ & $+3.6$ & $+3.9$ & $-6.8$ & $17{\to}5$ & $22{\to}22$ & 34 (9) \\
\rowcolor{gray!12}
GRSE     & $+7.4$ & $+6.0$ & $+4.4$ & $-3.1$ & $17{\to}7$ & $22{\to}14$ & 61 (5) \\
\bottomrule
\end{tabular}}
\caption{Answer transitions from the empty to the final library. Each test question is \emph{declined} (empty or ``not mentioned''), \emph{correct} (F1~$\geq 0.5$), or \emph{wrong} under each library. $\Delta$: change in mean token F1 ($\times$100). D$\to$C, W$\to$C, C$\to$W: contribution to $\Delta$ of questions moving from declined to correct, wrong to correct, and correct to wrong; the remaining transitions make up the rest. Decl.: share of declined answers, before $\to$ after. No ev.: share of questions whose retrieved evidence cites no dialogue turn (e.g., [D3:5]), before $\to$ after. \#D$\to$C: number of declined-to-correct questions, with those answered without cited evidence in (). No ev.\ and \#D$\to$C are reported for LoCoMo only, as HotpotQA has no retrieval step. All rows use the first run of each method; on LoCoMo, each category has its own library, and empty-library answers come from iteration~0 of the same GRSE run.}
\label{tab:transitions}
\end{table}

\paragraph{\emph{Where do the gains come from?}}
Table~\ref{tab:transitions} tracks each test question from the empty library to the final library. The methods differ mainly in how many correct answers they break: MASkills breaks three times as many as GRSE on HotpotQA with Qwen3.8 ($-1.8$ vs.\ $-0.6$) and twice as many on LoCoMo, with Gemma4 ($-6.8$ vs.\ $-3.1$), where its losses offset its fixes ($\Delta$: $-0.3$ vs.\ $+7.4$). With the Gemma4 actor on LoCoMo, declined questions that become correct account for most of GRSE's gain ($+6.0$ of $+7.4$), and the last two columns show how they are fixed. MASkills reduces declines without improving retrieval: the share of questions with no cited evidence stays at 22\%, and 9 of its 34 D $\to$ C lack evidence. GRSE lowers this share to 14\%, with only 5 of 61 lacking evidence, consistent with crediting Agent~1's search rather than Agent~2's answer.

\paragraph{\emph{What do the learned skills encode?}}
We label every final skill by its primary purpose with Qwen3.8-27B (Table~\ref{tab:skill-types}). On LoCoMo, retrieval procedures are the largest type for every method and actor. With Gemma4, GRSE also learns more abstention rules than MASkills (18\% vs.\ 5\%), in line with its declining rather than answering without evidence (Table~\ref{tab:transitions}). On HotpotQA with Qwen3.8, GRSE mostly writes answer-format rules (57\%), while MASkills mostly writes reasoning procedures (50\%). Together with its higher rate of broken answers, this suggests that MASkills' reasoning rules often redirect answers that were already correct. Appendix~\ref{sec:case-learned} walks through two examples: a Gemma4 retriever that learns to find the turn holding the answer, and a HotpotQA answer that GRSE's learned skills make it more specific.

\subsection{Robustness Analysis}
\label{sec:experiment-robustness}

A skill library may contain flawed or irrelevant skills, for example when skills are written by hand or reused from another task. We test whether skill evolution can repair such a library on LoCoMo single-hop questions, the category with the largest test set (453 questions), with the Qwen3.8 actor. Instead of an empty library, both methods start from the same corrupt library: each agent keeps one helpful skill and receives two harmful or irrelevant skills phrased as plausible advice, and neither the critic nor the optimizer is told which skills are perturbed. All other settings follow the per-category runs. We run GRSE and MASkills once each, and evaluate the corrupt library and its helpful skills alone as reference.

\begin{table}[t]
\centering
\small
\setlength{\tabcolsep}{3pt}
\begin{tabular}{lcccc}
\toprule
Library & F1 & BLEU & $\Delta$F1 & Abst.~(\%) \\
\midrule
Perturbed (start)   & 60.6 & 52.1 & --     & 9.7 \\
Helpful skills only & 66.7 & 57.4 & $+6.1$ & 4.2 \\
\midrule
MASkills            & 58.4 & 49.3 & $-2.2$ & 7.9 \\
\rowcolor{gray!12}
GRSE                & \textbf{66.9} & \textbf{57.4} & $\mathbf{+6.3}$ & 4.6 \\
\bottomrule
\end{tabular}
\caption{Repairing a perturbed skill library on LoCoMo single-hop questions (Qwen3.8 actor, 453 test questions, greedy decoding, one run per method). Token F1 and smoothed BLEU ($\times$100) are measured with the final library. $\Delta$F1 is the change in F1 from the perturbed starting library. Abst.\ is the percentage of refusal answers such as ``not mentioned''. Bold marks the best learned library. GRSE beats MASkills by $+8.5$ F1.}
\label{tab:robustness}
\end{table}

Table~\ref{tab:robustness} shows that GRSE recovers the score of the helpful skills alone ($66.9$ vs.\ $66.7$ F1) and halves the abstention rate ($9.7\to4.6$), whereas MASkills ends slightly below its starting point. Group-relative credit assignment attributes the damage to individual skills, so GRSE prunes or locally repairs the harmful ones. MASkills instead merges perturbed content into new skills. Appendix~\ref{sec:robustness-analysis} traces each corrupt skill and walks through individual cases.

\begin{table}[t]
\centering
\footnotesize
\setlength{\tabcolsep}{2.5pt}
\begin{tabular}{@{}llrlrl@{}}
\toprule
 & & \multicolumn{2}{c}{GRSE} & \multicolumn{2}{c}{MASkills} \\
\cmidrule(lr){3-4} \cmidrule(l){5-6}
Planted skill & Kind & \multicolumn{1}{c}{$u_1$} & Fate & \multicolumn{1}{c}{$u_1$} & Fate \\
\midrule
\multicolumn{6}{@{}l}{\textit{Agent~1 (retriever)}} \\
context expansion & help & $+0.18$ & merged\textsubscript{2}  & $+0.03$ & merged\textsubscript{2} \\
latest session    & harm & $-0.12$ & merged\textsubscript{2}  & \textcolor{red!75!black}{$+0.07$} & merged\textsubscript{2} \\
equation solving  & irr. & $0.00$  & kept                     & $0.00$  & merged\textsubscript{2} \\
\multicolumn{6}{@{}l}{\textit{Agent~2 (reasoner)}} \\
answer fidelity   & help & $+0.09$ & refined\textsubscript{1} & $+0.02$ & merged\textsubscript{4} \\
complete sentence & harm & $-0.22$ & refined\textsubscript{2} & $-0.12$ & merged\textsubscript{2} \\
verbatim guard    & harm & $-0.19$ & pruned\textsubscript{3}  & \textcolor{red!75!black}{$+0.04$} & merged\textsubscript{4} \\
\midrule
\multicolumn{2}{@{}l}{Correct sign (of 5)}                   & 5       & & 3       & \\
\multicolumn{2}{@{}l}{$\bar u_1$(help) $-$ $\bar u_1$(harm)} & $+0.31$ & & $+0.03$ & \\
\bottomrule
\end{tabular}
\caption{Credit assigned to the planted skills of the robustness study (Figure~\ref{fig:perturbed-skills}). Kind: helpful (help), harmful (harm), or irrelevant (irr.). $u_1$: aggregated utility after the first update; it has the correct sign if positive for helpful and negative for harmful skills (\textcolor{red!75!black}{red}: wrong sign). Fate: the skill's status by iteration~5, with the iteration of the change as subscript. Same runs as Table~\ref{tab:robustness}.}
\label{tab:credit-planted}
\end{table}
Table~\ref{tab:credit-planted} examines whether each method credits the planted skills correctly. For every planted skill, we read its utility after the first update ($u_1$), which aggregates the critic's credits over all training questions of iteration~1. A credit is correct if it is positive for a helpful skill and negative for a harmful one. GRSE assigns the correct sign to all five helpful and harmful skills and separates the two kinds by $0.31$, whereas MASkills gets two of the five wrong and separates them by only $0.03$. The clearest failure is Agent~2's harmful skill \texttt{verbatim-evidence-guard}: MASkills credits it positively and keeps raising its utility over the next two iterations, consistent with crediting a skill for successes in which it is merely present. GRSE instead marks it harmful and prunes it in iteration~3. Both methods assign the irrelevant skill zero utility, but GRSE leaves it untouched, whereas MASkills merges it into Agent~1's new retrieval protocol. This is the failure mode of trajectory-level credit that motivates group-relative comparison (\S\ref{sec:introduction}).

\subsection{Cross-Task Transfer}

\providecommand{\sd}[1]{\textsubscript{$\pm$#1}}
\begin{table}[t]
\centering
\small
\setlength{\tabcolsep}{5pt}
\begin{tabular}{@{}lcc@{}}
\toprule
Target & LoCoMo MH & HotpotQA \\
\midrule
Empty & 50.2 & 80.7 \\
\midrule
\multicolumn{3}{@{}l}{\textit{MASkills}} \\
\hspace{0.5em}In-domain & 46.7\sd{1.4} & 80.6\sd{0.6} \\
\hspace{0.5em}Transfer  & 50.2\sd{1.5} & 82.7\sd{0.7} \\
\midrule
\multicolumn{3}{@{}l}{\textit{GRSE}} \\
\rowcolor{gray!12}
\hspace{0.5em}In-domain & \textbf{53.4}\sd{0.8} & \textbf{83.7}\sd{0.5} \\
\rowcolor{gray!12}
\hspace{0.5em}Transfer  & 51.5\sd{0.6}  & 82.5\sd{0.0} \\
\bottomrule
\end{tabular}
\caption{Transfer between the two multi-hop tasks (Qwen3.8 actor, test F1 $\times$100, mean\,$\pm$\,SD over three runs). \emph{Transfer}: final libraries from the other task (HotpotQA libraries on LoCoMo multi-hop questions and vice versa); each agent keeps the target task's role. \emph{In-domain}: libraries learned on the target (Tables~\ref{tab:main-hotpotqa} and~\ref{tab:main-locomo}).}
\label{tab:transfer}
\end{table}
Do learned skills only fit the task they were learned on? HotpotQA and LoCoMo multi-hop questions require the same kind of reasoning over different sources, so we swap the final libraries between them (Qwen3.8 actor; Table~\ref{tab:transfer}). Transferred libraries of both methods are at least as good as the empty library, so the learned rules carry over across domains without harm. The methods differ in what training on the target adds. For GRSE, in-domain libraries beat transferred ones on both targets ($+1.9$ and $+1.2$ F1), yet the transferred libraries still improve over the empty library ($+1.3$ and $+1.8$), indicating that the skills learned on one task do not overfit to it. For MASkills, the in-domain libraries fall below the transferred ones ($-3.5$, $-2.1$) and do not exceed the empty library, consistent with the correct answers its updates break (\S\ref{sec:experiment:learned_skills}).

\subsection{Ablation studies}
\label{sec:experiment-ablation}

To isolate the contributions of structured diagnosis and group sampling, we learn one skill library shared by all four LoCoMo categories (Qwen3.8 actor, one run per setting, 842 test questions). We test four ablation settings: GRSE w/o structure; GRSE with $G{=}1$; GRSE with the DAPO filter; and GRSE w/ label (details see Appendix~\ref{app:ablation}). From Table~\ref{tab:ablation} we observe that: 
\paragraph{Both components help.} Group contrast gives the critic a basis for attribution, and structured analysis checks it. GRSE has the best category average in F1 (54.9) and BLEU (42.2), $+2.9$ F1 over the empty library, with the largest gains on temporal and open-domain questions. Removing structured diagnosis or group sampling ($G{=}1$) lowers the average F1 by 1.0 and 0.8 respectively. 
\paragraph{Equal-reward groups still inform the critic.} The DAPO filter discards the 7--13 of 32 groups per iteration in which all four rollouts receive the same reward. This lowers the category average by 0.5 F1 and 0.6 BLEU, mostly through open-domain questions ($-4.1$ F1), while single-hop questions gain ($+2.4$ F1). Even without contrast, equal-reward groups still show the critic which behaviors consistently fail or succeed. 
\paragraph{Labels favor the majority category.} GRSE w/ label tops the pooled result (All) only through single-hop questions, 54\% of the test set ($+4.7$ F1 and BLEU). It trails GRSE on every other category, most on open-domain ($-6.7$ F1), and has the lowest category average of all variants (53.0 F1). Label-conditioned evidence thus pulls a shared library toward the largest category, while degrading performance on the remaining ones.
\section{Conclusion}
\label{sec:conclusion}
We studied behavioral credit assignment in cooperative language-model teams whose agents keep fixed parameters and act through natural-language skills. GRSE executes each question several times under the same skill libraries, compares the executions agent by agent, and attributes the team's outcome to individual agent--skill pairs. Structured diagnosis grounds each attribution in checked execution steps, and the resulting credit drives targeted, validated skill updates. On HotpotQA and LoCoMo with two open-weight actors, GRSE outperforms MASkills on HotpotQA and every LoCoMo category except multi-hop, with the largest gains on multi-hop questions, where the answer depends on evidence passed between the two agents. Its gains also rest on better retrieval rather than on fewer declined answers alone. Accurate credit also makes GRSE robust: starting from a library with planted harmful skills, it credits every planted skill with the correct sign and restores the score of a clean library, whereas MASkills, which credits each trajectory on its own, rewards a harmful skill for successes in which it is merely present and ends below its starting point. Learned libraries also transfer between HotpotQA and LoCoMo multi-hop questions without harm, and only GRSE gains further from training on the target task itself. These results suggest that comparing executions of the same task is a practical way to determine which agent, and which of its behaviors, should change.

\section*{Limitations}
\label{sec:limitation}

GRSE trades training compute for more accurate credit: executing each question $G{=}4$ times raises actor-token cost by 1.6--2.9$\times$ over MASkills and wall-clock time by 1.3--1.8$\times$ (Appendix~\ref{sec:app:runtime}). The overhead is confined to training, since test-time inference is unchanged. Furthermore, GRSE is defined for any number of agents, but we evaluate two-agent sequential teams on two question-answering benchmarks with two open-weight actors and one critic model. Larger teams, non-sequential communication, longer tool-heavy tasks, and other critic models remain to be tested in the future.

\section*{Ethical Considerations}

In our experiments, we use only publicly available benchmarks, HotpotQA and LoCoMo, under their respective licenses. All models are open-weight and run locally. Because GRSE writes skills in natural language, humans can inspect and edit the learned libraries for safety before deployment.

\section*{Acknowledgments}

We used LLMs to help draft and restructure parts of the main text and appendices. The authors reviewed and revised all such material and take full responsibility for the methods, experimental results, claims, and references in the paper.
\bibliography{custom}
\appendix
\section{Experimental Details}
\label{sec:app:implementation}



\paragraph{Models and serving.} The actor is Qwen3.8-27B~\citep{qwen38} or Gemma4-31B-it~\citep{gemmateam2026gemma4}; Qwen3.8-27B serves as critic and optimizer in all runs. All models are served with vLLM~0.29 on two NVIDIA RTX A6000 GPUs each (tensor parallelism 2, context length 65{,}536), with thinking mode disabled. Generation is capped at 4{,}096 tokens for the actor, 4{,}000 for the critic, 1{,}200 for skill induction and consolidation, and 800 for refinement. The critic decodes in JSON mode with a structured diagnosis that violates the citation constraints is returned once for repair.

\paragraph{Data and checkpoints.} Table~\ref{tab:assets} lists the data, their splits, and model checkpoints. Both datasets are used in their official releases, and both models are loaded from Hugging Face. LoCoMo is split by conversation, so no conversation appears in two splits. The per-category setting trains one library per question category; the mixed setting trains one library on all four categories. Category 5 (adversarial) is excluded, following MASkills and Mem0~\citep{chhikara2025mem0}.

\begin{table}[t]
\centering
\footnotesize
\renewcommand{\arraystretch}{1.15}
\begin{tabular}{@{}>{\raggedright\arraybackslash}p{\columnwidth}@{}}
\toprule
\textit{Datasets} \\
\textbf{HotpotQA}: distractor setting, development set (all questions labeled \emph{hard}). Split: 100 training, 50 validation, and 500 test questions (indices 150--649). Source: \path{hotpot_dev_distractor_v1.json} from \url{https://hotpotqa.github.io}. License: CC~BY-SA~4.0. \\
\addlinespace
\textbf{LoCoMo}: 10 conversations, question categories 1--4. Split: 2 training, 2 validation, and 6 test conversations (seed 42, fixed for all runs); open-domain uses 2/2/5, since conv-30 has no open-domain question. Test questions: 156 multi-hop, 179 temporal, 52 open-domain, and 453 single-hop; 842 in Mix. Source: \path{locomo10.json} from \url{https://github.com/snap-research/locomo}. License: CC~BY-NC~4.0. \\
\midrule
\textit{Model checkpoints (Hugging Face)} \\
\textbf{Qwen3.8-27B}: \path{Qwen/Qwen3.8-27B}, revision \texttt{1d4bf0f}; actor, critic, and optimizer. \\
\textbf{Gemma4-31B}: \path{google/gemma-4-31B-it}, revision \texttt{842da37}; actor. \\
\bottomrule
\end{tabular}
\caption{Datasets, data splits, and model checkpoints.}
\label{tab:assets}
\end{table}

\paragraph{Decoding.} Training rollouts use each model's default sampling parameters (temperature 1.0, top-$p$ 0.95, top-$k$ 20 for Qwen and 64 for Gemma), so that rollouts of the same question can differ. On LoCoMo, validation is greedy; on HotpotQA, validation uses the training sampling parameters. All reported test scores evaluate the final (iteration-5) libraries with greedy decoding. MASkills on LoCoMo follows its released per-category runner, which decodes the actor greedily throughout.

\paragraph{Hyperparameters.} Table~\ref{tab:hyperparameters} lists the training settings. GRSE and MASkills share every setting except the number of rollouts per question: GRSE samples $G=4$ team trajectories per question and MASkills one. Both start from empty libraries and run five iterations. For skill evolution methods, we use the mean of category-specific token F1 and the smoothed BLEU as training reward on LoCoMo to reflect both answer content and exact wording; token F1 on HotpotQA. The operator schedule $\Omega$ applies refinement and induction every iteration, consolidation every second iteration, and pruning every second (HotpotQA) or third (LoCoMo) iteration. Both methods use the same output-format handling for skill files: induced and consolidated skills are normalized to valid front matter before parsing, and skill bodies may use up to 1{,}200 tokens.

\begin{table}[t]
\centering
\footnotesize
\setlength{\tabcolsep}{3pt}
\begin{tabular}{@{}lccc@{}}
\toprule
 & & \multicolumn{2}{c}{LoCoMo} \\
\cmidrule(l){3-4}
 & HotpotQA & Per Cat. & Mix \\
\midrule
Iterations                            & 5          & 5          & 5 \\
Batch size $|\mathcal{B}_t|$          & 100        & 24         & 32 \\
Rollouts per question $G$             & 4          & 4          & 4 \\
Validation set size $|\mathcal{V}|$   & 50         & 21--159    & 80 \\
Validation executions $H$             & 1          & 1          & 1 \\
Gate tolerance $\delta$               & 0.03       & 0.02       & 0.02 \\
Skill limit per agent $L$             & 8          & 4          & 8 \\
Advantage $\epsilon$                  & $10^{-8}$  & $10^{-8}$  & $10^{-8}$ \\
Training reward                       & F1         & \multicolumn{2}{c}{$\tfrac{1}{2}(\text{F1}+\text{BLEU})$} \\
\bottomrule
\end{tabular}
\caption{Training settings, shared by GRSE and MASkills except that MASkills uses one rollout per question. Per-category validation uses every question of that category in the two validation conversations (multi-hop 68, temporal 60, open-domain 21, single-hop 159).}
\label{tab:hyperparameters}
\end{table}

\paragraph{Baselines.} CoT-SC draws five samples at temperature 0.7 and returns the majority answer; Self-Refine runs at most three feedback rounds; all other prompting calls are greedy. The full-context baseline limits answers to 32 tokens.

\section{Learned Skills Analysis}

\label{sec:app:learned_skills}
This section explains the definitions behind \S\ref{sec:experiment:learned_skills}, the skill coding, and two cases. All analyses use the first run of each configuration and greedy decoding.

\paragraph{Answer states.} A final answer is \emph{declined} if it is empty or contains ``not mentioned'', ``not specified'', ``not available'', ``not found'', ``not stated'', ``no information'', ``unknown'' or ``cannot be determined’’. Otherwise it is \emph{correct} if its F1 score is at least 0.5 and \emph{wrong} if not. On LoCoMo, a question has \emph{evidence} if the retriever's output cites at least one dialogue turn (\texttt{[D}\textit{session}\texttt{:}\textit{turn}\texttt{]}). Outputs that say \texttt{NO\_EVIDENCE}, print a grep call instead of running it, or return nothing usable count as no evidence.

\paragraph{Skill coding.} Each skill in the final libraries (180 skills) is labeled by Qwen3.8-27B at temperature~0, given its name, description and body, with one primary type: \emph{format} (form of the final answer: length, wording, normalization), \emph{retrieval} (how to search for or expand evidence), \emph{verification} (checking that an answer is grounded in the evidence), \emph{reasoning} (a procedure for inferring the answer, such as temporal arithmetic or aggregation), \emph{abstention} (when to answer or decline, including rules that forbid declining), or \emph{generic} (advice without a concrete procedure). No skill was labeled generic. Table~\ref{tab:skill-types} gives the shares.

\begin{table}[t]
\centering
\footnotesize
\setlength{\tabcolsep}{2.8pt}
\begin{tabular}{@{}lrrrrrrr@{}}
\toprule
 & & \multicolumn{5}{c}{Share of skills (\%)} & \\
\cmidrule(lr){3-7}
Method & $n$ & Fmt & Retr & Verif & Reas & Abst & Words \\
\midrule
\multicolumn{8}{@{}l}{\textit{HotpotQA, Qwen3.8}} \\
MASkills         & 6 & 17 &  0 & 33 & 50 & 0 & 504 \\
GRSE w/o struct. & 8 & 50 &  0 & 25 & 25 & 0 & 606 \\
GRSE w/ struct.  & 7 & 57 &  0 & 29 & 14 & 0 & 502 \\
\multicolumn{8}{@{}l}{\textit{HotpotQA, Gemma4}} \\
MASkills         & 5 & 60 &  0 & 20 & 20 & 0 & 507 \\
GRSE w/o struct. & 5 & 40 & 20 & 20 & 20 & 0 & 534 \\
GRSE w/ struct.  & 6 & 33 & 17 & 33 & 17 & 0 & 459 \\
\midrule
\multicolumn{8}{@{}l}{\textit{LoCoMo, Qwen3.8}} \\
MASkills         & 23 & 26 & 48 & 13 & 13 &  0 & 439 \\
GRSE w/o struct. & 26 & 27 & 50 &  4 & 19 &  0 & 465 \\
GRSE w/ struct.  & 30 & 37 & 47 &  3 & 13 &  0 & 390 \\
\multicolumn{8}{@{}l}{\textit{LoCoMo, Gemma4}} \\
MASkills         & 20 & 25 & 50 &  5 & 15 & 5 & 403 \\
GRSE w/o struct. & 27 & 30 & 44 &  7 & 11 & 7 & 424 \\
GRSE w/ struct.  & 17 & 24 & 53 &  0 & 6 & 18 & 370 \\
\bottomrule
\end{tabular}
\caption{Final skill libraries by primary skill type. Each skill is assigned one primary type by Qwen3.8-27B: answer formatting (Fmt), retrieval (Retr), verification (Verif), reasoning (Reas), or abstention (Abst). $n$: number of skills. Words: mean length of description and body. On LoCoMo, the four per-category libraries are pooled. w/ and w/o struct.: with and without structured failure analysis. First runs.}
\label{tab:skill-types}
\end{table}

\paragraph{Learned Skill Cases.}
\label{sec:case-learned}

\begin{figure*}[t]
\centering
\begin{promptpanel}{Question: What type of pizza is James' favorite? \quad Gold: Pepperoni}
\begin{tcbraster}[promptraster, raster equal height=rows]
\begin{promptpanel}[colbacktitle=red!12]{Empty library}
\phead{Agent 1 search}
\texttt{pizza} $\rightarrow$ [D9:16], [D9:17], [D9:19], [D9:20] (not [D9:18], which lacks the keyword)\\
\texttt{favorite pizza}, \texttt{James said: ".*pizza} $\rightarrow$ no match
\phead{Agent 1 evidence}
\texttt{NO\_EVIDENCE}
\phead{Agent 2 answer}
\texttt{not mentioned} \hfill Score 0.0
\end{promptpanel}
\begin{promptpanel}[colbacktitle=green!12]{GRSE w/o structure (final)}
\phead{Agent 1 search}
\texttt{pizza|favorite} $\rightarrow$ [D9:17] is a question $\rightarrow$ \texttt{D9:17|D9:18|D9:19|D9:20}
\phead{Agent 1 evidence}
{[D9:17]} John: ``Pizza? Cool, I love pizza too! Which one do you love the most?''\\
{[D9:18]} James: ``Pepperoni of course! An amazing combination of spicy salami and cheese.''
\phead{Agent 2 answer}
\texttt{Pepperoni} \hfill Score 1.0
\phead{Active retrieval skill}
\texttt{adjacent-context-\allowbreak expansion}: ``If the retrieved turn is clearly a question \ldots, fetch the next turn [Dx:y+1].'' ``Answers often reside in the response turn immediately following the question or topic introduction.''
\end{promptpanel}
\end{tcbraster}
\end{promptpanel}
\caption{Learned Skill Case~A: a declined question becomes correct with cited evidence (Gemma4 actor, LoCoMo single-hop, GRSE w/o structure). The answer turn does not contain the keyword of the question; the learned skill fetches the turn that follows a matched question. Evidence abridged.}
\label{fig:case-learned-gemma}
\end{figure*}

\begin{figure*}[t]
\centering
\begin{promptpanel}{Question: What occupation was shared by David Yates and Pietro Germi? \quad Gold: director}
\begin{tcbraster}[promptraster, raster equal height=rows]
\begin{promptpanel}[colbacktitle=red!12]{Empty library}
\phead{Agent 1 output}
Filmmaker (or director).
\phead{Agent 2 answer}
\texttt{Filmmaker} \hfill Score 0.0
\end{promptpanel}
\begin{promptpanel}[colbacktitle=green!12]{GRSE w/o structure (final)}
\phead{Agent 1 output (abridged)}
\textbf{Classify Question Type}: Entity/Attribute Question. The question asks for a shared ``occupation'' (a role/attribute) \ldots\\
\textbf{Constraint-Based Entity Validation \& Grounding}: David Yates \ldots\ ``English \textbf{filmmaker} who has directed feature films''; Pietro Germi \ldots\ ``actor, screenwriter, and \textbf{director}.'' \ldots\\
The intersection of their explicitly stated occupations is \textbf{director} \ldots\ ``Filmmaker'' is a broader term \ldots
\phead{Agent 2 answer}
\texttt{director} \hfill Score 1.0
\phead{Learned Agent 1 skills used}
\texttt{answer-disambiguation-and-\allowbreak minimalism}: ``Classify Question Type: Explicitly categorize the question before intent identification. \ldots\ Attribute Question \ldots\ Target must be the specific value \ldots''\\
\texttt{constraint-based-entity-\allowbreak validation}: ``Extract Constraints: Identify all explicit constraints in the question. These include specific roles \ldots''
\end{promptpanel}
\end{tcbraster}
\end{promptpanel}
\caption{Learned Skill Case~B: a wrong answer becomes correct (Qwen3.8 actor, HotpotQA, GRSE w/o structure). The final Agent~1 follows the steps of its learned skills, whose headings it reproduces.}
\label{fig:case-learned-hotpotqa}
\end{figure*}

The two cases below illustrate the mechanisms of \S\ref{sec:experiment:learned_skills}. Both use the first run of each configuration and greedy decoding.

\textbf{\emph{Case A: fetching the answer turn (Gemma4, LoCoMo single-hop, GRSE).}} On \emph{What type of pizza is James' favorite?} (gold: \emph{Pepperoni}), the answer turn does not contain the keyword of the question: John asks ``Which one do you love the most?'' and James replies ``Pepperoni of course!'' (Figure~\ref{fig:case-learned-gemma}). With the empty library, the retriever greps \texttt{pizza}, which matches the turns around the answer but not the answer itself, tries two narrower patterns that match nothing, and returns \texttt{NO\_EVIDENCE}, so the reasoner declines. With the final library, the retriever recognizes the matched turn as a question and fetches the turns that follow it, as its learned skill \texttt{adjacent-context-expansion} prescribes, and the reasoner answers correctly. This question is one of the 56 declined questions that GRSE turns correct with cited evidence (\S\ref{sec:experiment:learned_skills}), and the responsible skill is a retrieval procedure, the most common skill type on LoCoMo (Table~\ref{tab:skill-types}). The skill performs the same question--answer expansion that repairs the perturbed library in Robustness Case~1 (Appendix~\ref{sec:robustness-analysis}, Figure~\ref{fig:case-retrieval}).

\textbf{\emph{Case B: a more specific answer (Qwen3.8, HotpotQA, GRSE).}} On \emph{What occupation was shared by David Yates and Pietro Germi?} (gold: \emph{director}), the passages call Yates a ``filmmaker who has directed'' films and Germi an ``actor, screenwriter, and director''. With the empty library, Agent~1 offers ``Filmmaker (or director)'' and Agent~2 keeps the broader term (Figure~\ref{fig:case-learned-hotpotqa}). With the final library, Agent~1 follows the steps of two learned skills, \texttt{answer-disambiguation-and-minimalism} and \texttt{constraint-based-entity-validation}: it classifies the question as asking for an attribute, checks each person's stated roles against the passages, and keeps the role both passages support. Agent~2 answers ``director''. This is the typical HotpotQA gain in Table~\ref{tab:transitions}: a wrong answer becomes correct without new evidence, through a more precise reading of the passages that the empty library already had.

\section{Robustness Analysis}
\label{sec:robustness-analysis}
\begin{figure*}[t]
\centering
\begin{tcbraster}[promptraster]
\begin{promptpanel}{Agent 1 (harmful)}
\textit{Name:} \texttt{focused-latest-session-retrieval}\\
\textit{Description:} Use when retrieving evidence: keeps the grep budget small and avoids stale, superseded facts.

\phead{Focused Retrieval Protocol}
Long conversations contain outdated statements; the most recent session reflects the speaker's current state.
\begin{enumerate}
  \item Pick the single most distinctive keyword from the question (one word, no alternation). Example: for ``What instrument does Sam play?'' use \texttt{<grep>instrument</grep>}.
  \item Issue exactly ONE grep. Additional searches waste budget and pull in contradictory older turns.
  \item From the hits, keep only lines from the highest-numbered session; discard all earlier sessions as outdated.
  \item Hand off at most two evidence lines.
\end{enumerate}
\phead{Do NOT emit}
\begin{itemize}
  \item more than one \texttt{<grep>} call
  \item evidence lines from sessions older than the latest hit
\end{itemize}
\end{promptpanel}
\begin{promptpanel}{Agent 1 (irrelevant)}
\textit{Name:} \texttt{algebraic-equation-solving}\\
\textit{Description:} Use when a question involves solving linear or quadratic equations for an unknown.

\phead{Equation Solving Procedure}
\begin{enumerate}
  \item Move all terms containing the unknown to the left-hand side and constants to the right.
  \item For linear equations $ax + b = c$, compute $x = (c - b) / a$.
  \item For quadratics $ax^2 + bx + c = 0$, compute the discriminant $d = b^2 - 4ac$; if $d \geq 0$, return $x = (-b \pm \sqrt{d}) / (2a)$.
  \item Substitute each root back into the original equation to verify it.
\end{enumerate}
\phead{Do NOT emit}
\begin{itemize}
  \item roots that fail the substitution check
\end{itemize}
\end{promptpanel}
\begin{promptpanel}{Agent 2 (harmful)}
\textit{Name:} \texttt{complete-sentence-answers}\\
\textit{Description:} Use when writing the final answer: graders reward complete, self-contained answers.

\phead{Complete Answer Format}
A bare phrase is ambiguous out of context. Always answer in one full sentence that restates the question's subject and verb, then add one clause of supporting context from the evidence.
\begin{itemize}
  \item Q: What did Sam start learning? $\rightarrow$ \texttt{Sam started learning the violin because a friend recommended it last spring.}
  \item Q: Where did Alex move? $\rightarrow$ \texttt{Alex moved to Denver after accepting a new job offer there.}
\end{itemize}
\phead{Do NOT emit}
\begin{itemize}
  \item one- or two-word answers
  \item answers that do not restate the subject of the question
\end{itemize}
\end{promptpanel}
\begin{promptpanel}{Agent 2 (harmful)}
\textit{Name:} \texttt{verbatim-evidence-guard}\\
\textit{Description:} Use when deciding whether to answer: hallucinated answers are penalized more than abstentions.

\phead{Verbatim Evidence Guard}
Answer only if the retriever evidence contains the answer verbatim as one contiguous span.
\begin{enumerate}
  \item If the answer requires any paraphrase, inference, or combining two turns, output exactly \texttt{not mentioned}.
  \item If the evidence mentions the topic but not the exact requested attribute, output exactly \texttt{not mentioned}.
  \item Never guess from partial or indirect evidence.
\end{enumerate}
\begin{itemize}
  \item Evidence: \texttt{Sam said, "I've been practicing every night since I got my new guitar."} Q: What instrument does Sam play? $\rightarrow$ \texttt{not mentioned} (the word ``plays'' never appears).
\end{itemize}
\phead{Do NOT emit}
\begin{itemize}
  \item answers inferred from indirect evidence
\end{itemize}
\end{promptpanel}
\begin{promptpanel}[raster multicolumn=2]{Helpful skills (one per agent, taken from a GRSE library trained on single-hop questions)}
Agent~1, \texttt{multi-hop-context-expansion}: ``Forces the retriever to expand grep results to adjacent turns to capture Q\&A pairs, session details, and antecedents required for complete answers.''\\
Agent~2, \texttt{answer-fidelity-and-temporal-precision}: ``Enforces strict causal isolation, temporal constraint validation, and preservation of specific activity nouns in final answers.''\\
Their full bodies (537 and 341 words) are omitted for space.
\end{promptpanel}
\end{tcbraster}
\caption{The perturbed seed library of the robustness experiment (\S\ref{sec:experiment-robustness}). Each agent starts with three skills: one helpful skill and two harmful or irrelevant ones. Skill bodies are reproduced verbatim; labels are for the reader only.}
\label{fig:perturbed-skills}
\end{figure*}

This subsection details the robustness experiment of \S\ref{sec:experiment-robustness}: the perturbed library and two cases (Cases~1 and~2).

\paragraph{Setup and perturbed library.} For robustness analysis, both GRSE and MASkills start from the same seed library of three skills per agent instead of an empty library (Figure~\ref{fig:perturbed-skills}). The library limit is four skills per agent, so one induction slot stays free. Each agent keeps one helpful skill, copied verbatim from the final library of the first GRSE analysis run on LoCoMo single-hop questions. The other skills are harmful or irrelevant. Harmful skills are phrased as plausible advice and use invented names, never entities from the test conversations. The agents, the critic, and the optimizer see each skill's name, description, and body, without knowing which one is \emph{helpful}, \emph{harmful}, and \emph{irrelevant} as shown in Figure~\ref{fig:perturbed-skills}. All other settings follow the per-category runs (24 training questions per iteration, five iterations, $\delta=0.02$, $L=4$), and each method is trained once. All scores are greedy test scores on the 453 single-hop test questions.

\paragraph{Robustness Cases} The following cases show how GRSE repairs harmful or irrelevant skills on LoCoMo single-hop questions. 


\begin{figure*}[htbp]
\centering
\begin{promptpanel}{Question: What is the name of John's cousin's dog? \quad Gold: Luna}
\begin{tcbraster}[promptraster, raster columns=3, raster equal height=rows]
\begin{promptpanel}[colbacktitle=red!12]{Perturbed library (start)}
\phead{Agent 1 evidence}
{[D31:20]} John: ``By the way, here is my cousin's dog.'' [image: a dog sitting on a rug]
\phead{Agent 2 answer}
\texttt{not mentioned} \hfill Score 0.0
\phead{Active retrieval rule}
\texttt{focused-latest-session-\allowbreak retrieval}: ``Pick the single most distinctive keyword \ldots\ Issue exactly ONE grep \ldots\ Hand off at most two evidence lines.''
\end{promptpanel}
\begin{promptpanel}[colbacktitle=red!12]{MASkills (final)}
\phead{Agent 1 evidence}
{[D31:20]} John: ``By the way, here is my cousin's dog.''
\phead{Agent 2 answer}
\texttt{not mentioned} \hfill Score 0.0
\phead{Active retrieval rule}
\texttt{comprehensive-query-\allowbreak protocol} (``Unified protocol for solving algebraic equations and \ldots\ retrieval''): ``If NEITHER multi-entity nor causal conditions are met \ldots, use standard single-keyword retrieval. Do not apply heuristics to single-entity, single-hop facts.''
\end{promptpanel}
\begin{promptpanel}[colbacktitle=green!12]{GRSE (final)}
\phead{Agent 1 evidence}
{[D31:20]} John: ``By the way, here is my cousin's dog.''\\
{[D31:21]} James: ``This pup is so adorable! What's their name?''\\
{[D31:22]} John: ``Their name is Luna.''
\phead{Agent 2 answer}
\texttt{Luna} \hfill Score 1.0
\phead{Active retrieval rule}
\texttt{retrieval-precision-\allowbreak expansion}: ``When a grep hit contains a direct question (e.g., `What is the name\ldots'), the answer is almost always in the next turn. \ldots\ Do not output the question turn as final evidence.''
\end{promptpanel}
\end{tcbraster}
\end{promptpanel}
\caption{Robustness Case~1: the harmful skill \texttt{focused-latest-session-retrieval}. The same single-hop test question is answered with the perturbed library and with the final libraries of MASkills and GRSE (greedy decoding; evidence abridged to the cited turns). MASkills keeps the single-keyword behavior of the harmful skill. GRSE keeps the question--answer expansion of the helpful one.}
\label{fig:case-retrieval}
\end{figure*} 

\textbf{\emph{Case 1: the harmful retrieval rule (Figure~\ref{fig:case-retrieval}).}} The perturbed Agent~1 skill \texttt{focused-latest-session-retrieval} tells the retriever to issue a single one-keyword search, keep only the latest session, and hand off at most two evidence lines. On \emph{What is the name of John's cousin's dog?} (gold: \emph{Luna}), this rule makes the perturbed library stop at the turn that introduces the dog, leaving the name never been retrieved, and Agent~2 answers ``not mentioned''. GRSE merged this skill with the helpful skill \texttt{multi-hop-context-expansion}, and the merged skill keeps the instruction to fetch the turn that answers a retrieved question, so the retriever returns the question--answer pair and Agent~2 answers ``Luna''. MASkills merged the same skills, together with the irrelevant algebra procedure, into one protocol that falls back to ``standard single-keyword retrieval'' for single-hop facts. Its retriever returns exactly the evidence of the perturbed library and fails in the same way. The same pattern decides other single-hop questions, for example \emph{How will John share the honey garlic chicken recipe?} and \emph{Will there be an interview required to volunteer?}, where the perturbed library and MASkills retrieve only the turn that asks.


\begin{figure*}[htbp]
\centering
\begin{tcbraster}[promptraster]
\begin{promptpanel}[colbacktitle=green!12]{GRSE: group-relative credit}
\phead{Iteration 1}
Critic, comparing rollouts of one question: ``In rollouts 0 and 3, agent\_2 received the correct evidence turns \ldots\ but output `not mentioned'. The skill's rule `If the answer requires any paraphrase, inference, or combining two turns, output exactly not mentioned' caused the agent to fail \ldots'' \hfill utility $-0.19$
\phead{Iteration 2}
Labeled harmful again \hfill utility $-0.07$
\phead{Iteration 3}
Pruned (harmful in 7 question groups over iterations 1--3).
\phead{Test (final library)}
Abstention 9.7\% $\rightarrow$ 4.6\%
\end{promptpanel}
\begin{promptpanel}[colbacktitle=red!12]{MASkills: per-trajectory credit}
\phead{Iterations 1--3}
Refined each iteration \hfill utility $0.04 \rightarrow 0.09 \rightarrow 0.12$\\
Kept: ``Q: What instrument does Sam play? $\rightarrow$ \texttt{not mentioned} (the word `plays' never appears).''
\phead{Iteration 4}
Merged with the helpful answer skill into \texttt{answer-precision-and-fidelity}: ``Answer only if evidence contains the answer as a contiguous span or minor syntactic variant.''
\phead{Test (final library)}
Abstention 9.7\% $\rightarrow$ 7.9\%; answers 6.6 words on average
\end{promptpanel}
\end{tcbraster}
\caption{Robustness Case~2: the harmful skill \texttt{verbatim-evidence-guard} during training. Utility is the library's running estimate of the skill's value, aggregated from critic credits.}
\label{fig:case-abstention}
\end{figure*}

\textbf{\emph{Case 2: the harmful abstention rule (Figure~\ref{fig:case-abstention}).}} The perturbed skill \texttt{verbatim-evidence-guard} tells Agent~2 to answer only from a verbatim contiguous span and otherwise to output ``not mentioned''. GRSE's critic compares rollouts of the same question and finds the skill active in rollouts that had the correct evidence but answered ``not mentioned''. It labeled the skill harmful in seven question groups, its utility estimate stayed negative ($-0.19$ after iteration~1, $-0.07$ after iteration~2), and pruning removed it in iteration~3. MASkills credits each trajectory on its own, without a contrast between rollouts of the same question. Its utility estimate for the skill instead grew ($0.04$, $0.09$, $0.12$ after iterations~1--3), consistent with crediting a skill that is present in every trajectory, including the successful ones, and MASkills kept refining the skill instead of removing it. The refined skill still carried the original example that answers ``not mentioned'' when the evidence does not repeat the question's verb. In iteration~4, consolidation merged it with the helpful answer skill, and the merged skill still answers ``only if evidence contains the answer as a contiguous span or minor syntactic variant''.

\section{Ablation Analysis}
\label{app:ablation}
\begin{table*}[t]
\centering
\small
\setlength{\tabcolsep}{4pt}
\begin{tabular}{@{}lcccccccccccc@{}}
\toprule
 & \multicolumn{2}{c}{Multi-hop} & \multicolumn{2}{c}{Temporal} & \multicolumn{2}{c}{Open-domain} & \multicolumn{2}{c}{Single-hop} & \multicolumn{2}{c}{All} & \multicolumn{2}{c}{Avg.} \\
\cmidrule(lr){2-3} \cmidrule(lr){4-5} \cmidrule(lr){6-7} \cmidrule(lr){8-9} \cmidrule(lr){10-11} \cmidrule(l){12-13}
Setting & F1 & BLEU & F1 & BLEU & F1 & BLEU & F1 & BLEU & F1 & BLEU & F1 & BLEU \\
\midrule
Empty library (iter.~0) & 50.9 & 43.5 & 70.8 & 48.2 & 26.8 & 16.7 & 59.6 & 50.8 & 58.3 & 46.7 & 52.0 & 39.8 \\
\midrule
\rowcolor{gray!12}
GRSE & 52.3 & \underline{44.0} & \textbf{73.1} & \textbf{51.5} & \textbf{33.9} & \textbf{22.3} & 60.2 & 50.9 & 59.8 & 47.9 & \textbf{54.9} & \textbf{42.2} \\
\quad w/o structure & 51.5 & 42.1 & 70.6 & 48.5 & \underline{31.6} & \underline{20.9} & 62.0 & 52.4 & 59.9 & 47.7 & 53.9 & 41.0 \\
\quad $G{=}1$ & \textbf{53.7} & \textbf{44.6} & \underline{72.0} & \underline{51.3} & 30.5 & 19.4 & 60.4 & 51.4 & 59.7 & 48.1 & 54.1 & \underline{41.7} \\
\quad DAPO filter & \underline{53.4} & \underline{44.0} & 71.6 & 51.0 & 29.8 & 18.1 & \underline{62.6} & \underline{53.4} & \underline{60.7} & \underline{48.9} & \underline{54.4} & 41.6 \\
\quad w/ label & 49.2 & 40.4 & 70.5 & 49.8 & 27.2 & 18.7 & \textbf{64.9} & \textbf{55.6} & \textbf{60.8} & \textbf{49.2} & 53.0 & 41.1 \\
\bottomrule
\end{tabular}
\caption{Ablation on LoCoMo with one shared library learned on all four categories together (Qwen3.8 actor, greedy decoding, one run per setting, 842 test questions). Token F1 and smoothed BLEU ($\times$100) per question category, over all questions (All), and averaged over the four categories with equal weight (Avg.). Single-hop questions make up 54\% of All. Best per column in \textbf{bold}, second best \underline{underlined}. \emph{w/o structure} removes the failure-diagnosis schema and its checks. \emph{$G{=}1$} samples one team trajectory per question instead of four, so no group-relative comparison is possible. \emph{DAPO filter} keeps four rollouts but, following the dynamic sampling of DAPO, sends only groups whose rewards differ to the critic. \emph{w/ label} conditions the critic on the question category and groups update evidence by category.}
\label{tab:ablation}
\end{table*}

Table~\ref{tab:ablation} shows the ablation results. GRSE w/o structure removes the failure-diagnosis schema and its checks. GRSE with $G{=}1$ samples one trajectory per question, leaving the critic no in-group contrast. GRSE with the DAPO filter keeps four rollouts but, following the dynamic sampling of DAPO~\citep{yu2026dapo}, passes only groups with differing rewards to the critic. GRSE w/ label, which we assume different tasks should have unique skills, additionally conditions the critic on the question category and groups update evidence by category.

\section{Runtime Analysis}
\label{sec:app:runtime}

\providecommand{\sd}[1]{\textsubscript{$\pm$#1}}
\begin{table}[t]
\centering
\footnotesize
\setlength{\tabcolsep}{4pt}
\begin{tabular}{@{}lcccc@{}}
\toprule
 & \multicolumn{2}{c}{HotpotQA} & \multicolumn{2}{c}{LoCoMo} \\
\cmidrule(lr){2-3} \cmidrule(l){4-5}
 & Qwen3.8 & Gemma4 & Qwen3.8 & Gemma4 \\
\midrule
\multicolumn{5}{@{}l}{\textit{Wall-clock training time (h)}} \\
MASkills & 2.8\sd{0.8} & 4.6\sd{1.1} & 15.9\sd{0.7} & 19.0\sd{1.3} \\
GRSE     & 4.3\sd{0.6} & 5.9\sd{0.6} & 20.8\sd{1.6} & 23.4\sd{2.5} \\
\multicolumn{5}{@{}l}{\textit{Actor tokens, training rollouts (M)}} \\
MASkills & 2.7\sd{0.3}  & 2.6\sd{0.2}  & 6.1\sd{0.2}  & 4.2\sd{0.3} \\
GRSE     & 10.9\sd{1.5} & 11.3\sd{0.3} & 23.3\sd{1.1} & 17.6\sd{0.5} \\
\multicolumn{5}{@{}l}{\textit{Actor tokens, training + validation (M)}} \\
MASkills & 4.4\sd{0.5}  & 4.3\sd{0.4}  & 26.6\sd{0.8} & 19.4\sd{1.1} \\
GRSE     & 12.6\sd{1.8} & 13.0\sd{0.3} & 45.4\sd{2.6} & 35.4\sd{1.0} \\
\bottomrule
\end{tabular}
\caption{Training cost of the main-result runs (mean\,$\pm$\,SD over three runs of five iterations; LoCoMo sums the four per-category runs). Actor tokens are the input and output tokens of both agents; critic and optimizer tokens are not logged and are excluded.}
\label{tab:runtime}
\end{table}

Table~\ref{tab:runtime} compares the training cost of GRSE and MASkills for the main-result runs.

\paragraph{Token usage.} GRSE executes each training question $G{=}4$ times, so its training rollouts use 3.8--4.3$\times$ the actor tokens of MASkills (e.g., 23.3M vs.\ 6.1M on LoCoMo with Qwen3.8). Validation costs about the same for both methods, which reduces the overall overhead to 1.7--3.0$\times$ (45.4M vs.\ 26.6M). Both methods call the critic equally often, but each GRSE call reads all $G$ executions and is longer.

\paragraph{Wall-clock time.} Training time grows by only 1.2--1.5$\times$, because the executions of a group run in parallel. Times are approximate, since some runs shared servers with other jobs.
\section{Prompt Templates}
\label{sec:app:template}

This appendix lists the prompts used in our experiments (Figures~\ref{fig:prompt-hotpotqa-agents}--\ref{fig:prompt-operators}). Placeholders such as \ph{question} are filled in at run time. The agent roles and the LoCoMo operator prompts follow MASkills~\citep{yao2026maskills}. The LoCoMo full-context baseline follows the benchmark's own QA protocol~\citep{maharana2024evaluating}.

\paragraph{Agent roles.} Each agent's system prompt is its fixed role followed by its current skill library: the role, a heading \texttt{\#\# Skills}, and the bodies of all skills in the library. Roles do not change during training. On HotpotQA (Figure~\ref{fig:prompt-hotpotqa-agents}), the two agents act in sequence: Agent~1 sees the question and passages, and Agent~2 additionally sees Agent~1's response and gives the final answer. On LoCoMo (Figures~\ref{fig:prompt-locomo-agent1} and~\ref{fig:prompt-locomo-agent2}), Agent~1 searches the conversation with \texttt{<grep>} calls (at most three, each returning up to 20 matching turns), and Agent~2 answers from the retrieved evidence only. The environment appends the question, its category, and the tool-call instructions to these roles.

\paragraph{Single-model baselines.} These baselines send one user message per call, with no system prompt, agents, or skills (Figure~\ref{fig:prompt-baselines}). On HotpotQA, \ph{context} holds the same ten distractor-setting passages that the agents see. CoT-SC samples the CoT prompt five times and returns the majority answer. Self-Refine starts from the CoT answer, then alternates feedback and refinement until the feedback is \texttt{CORRECT} or three rounds have passed. The LoCoMo full-context baseline follows the benchmark's own QA protocol~\citep{maharana2024evaluating}: \ph{conversation} is the whole conversation, with each session rendered as \texttt{DATE: \{date\}}, then \texttt{CONVERSATION:}, then one line per turn, \texttt{\{speaker\} said, "\{text\}"}. Answers are limited to 32 tokens.

\paragraph{Group-relative credit assignment.} For each training question, the critic receives the instructions in Figure~\ref{fig:prompt-critic}, followed by \texttt{INPUT:} and a JSON payload. The payload holds the task (question and passages, or the LoCoMo question without the conversation), each agent's current skill library, and all rollouts of the group with their reward, group-relative advantage, agent steps, and recorded skill use. Repeated text (the passages inside each step, identical system prompts) is replaced by references to keep the payload within the context window. The critic decodes in JSON mode. On HotpotQA, an answer-length note is appended.

\paragraph{Structured failure diagnosis.} GRSE with structured analysis appends the instructions in Figure~\ref{fig:prompt-structured} to the critic prompt and adds two fields to the payload: \texttt{citation\_constraints}, which lists the legal full- and non-full-reward rollout indices and the agent of each step, and \texttt{task\_category}. Without the category label, \texttt{task\_category} is \texttt{not\_grouped}; with it, \texttt{task\_category} is the question category for LoCoMo. Diagnoses that violate the constraints are returned to the critic once for repair.

\paragraph{Skill update operators.} Induction creates a new skill from the induction evidence collected for an agent, refinement applies anchored edits to an existing skill, and consolidation merges overlapping skills (Figure~\ref{fig:prompt-operators}). Induction and consolidation end with an output skeleton so that every new skill is a well-formed skill file. On LoCoMo, the operators use category-specific prompts followed by the output-format note.

\begin{figure}[t]
\centering
\begin{promptpanel}{Agent 1}
You are participating in a collaborative task with 2 agents.
\begin{itemize}
  \item You are Agent 1, speaking first
  \item After you respond, Agent 2 will see the task and your response
  \item Agent 2's output will be the final answer
\end{itemize}
Please provide your response to the task.
\end{promptpanel}

\vspace{4pt}
\begin{promptpanel}{Agent 2}
You are participating in a collaborative task with 2 agents.
\begin{itemize}
  \item You are Agent 2, speaking last
  \item You can see the original task and the responses from Agent 1
  \item YOUR output is the FINAL answer that will be evaluated
  \item For HotPotQA, your FINAL answer must be the shortest faithful form of the answer (an entity name, a year, or a short phrase). Output ONLY that answer string -- do NOT include reasoning, explanations, restatements of the question, citations, or surrounding sentences.
\end{itemize}
Please provide your response to the task.
\end{promptpanel}
\caption{Agent roles on HotpotQA.}
\label{fig:prompt-hotpotqa-agents}
\end{figure}

\begin{figure}[t]
\centering
\begin{promptpanel}{Agent 1 (retriever)}
You are the \textbf{retriever} in a multi-agent QA team answering questions about a long multi-session conversation between two speakers (the LOCOMO benchmark).

\phead{Inputs you receive}
\begin{itemize}
  \item The question and a category hint (1 multi-hop, 2 temporal, 3/4 single-hop).
  \item The conversation, rendered oldest $\rightarrow$ newest, possibly truncated to fit context. Every turn is prefixed with its dialog id, e.g.\ \texttt{[D3:7] Caroline said, "..."}. Session headers carry the calendar date, e.g.\ \texttt{-{}-{}- session\_3 (2023-06-19) -{}-{}-}.
\end{itemize}

\phead{What you must output}
\begin{itemize}
  \item Up to 6 short verbatim excerpts that bear on the question, each on its own line, in the form: \texttt{D<sess>:<turn> SPEAKER said: "<short verbatim text>"}.
  \item Prefer minimal, citable evidence over paraphrase. Keep each excerpt under \textasciitilde 25 words.
  \item For temporal (cat 2) questions, also include the relevant session date when known.
\end{itemize}

\phead{Hard constraints}
\begin{itemize}
  \item Do NOT answer the question. Your output is consumed by the downstream reasoner; an opinion or guess from you will pollute its evidence pool.
  \item Do NOT invent dia\_ids -- only cite IDs that appear in the conversation.
\end{itemize}
\end{promptpanel}
\caption{Agent~1 role on LoCoMo.}
\label{fig:prompt-locomo-agent1}
\end{figure}

\begin{figure*}[t]
\centering
\begin{promptpanel}[sidebyside, fontupper=\scriptsize, fontlower=\scriptsize]{Agent 2 (reasoner)}
You are the \textbf{reasoner} in a multi-agent QA team for the LOCOMO benchmark.

\phead{Inputs you receive}
\begin{itemize}
  \item The question and a category hint.
  \item The retriever's evidence excerpts (and any earlier agents' messages).
  \item You do NOT see the full conversation directly -- work from the cited evidence.
\end{itemize}

\phead{Output rules (LOCOMO grader formatting -- read carefully)}
\begin{itemize}
  \item Output ONLY the final answer. No reasoning, no preface, no quotes around it, no \texttt{"Final answer:"} prefix.
  \item \textbf{Single-answer categories (2, 3, 4)}: output exactly one value.
  \begin{itemize}
    \item \textbf{Category 2} (temporal): a SINGLE date or year, taken from the cited evidence. Prefer the format used by the conversation (e.g.\ \texttt{7 May 2023}, \texttt{2023}). If the cited turn does not carry an explicit date, use its session date. Do NOT list multiple candidates separated by commas; pick one.
    \begin{itemize}
      \item \textbf{NEVER emit ISO format} like \texttt{2023-05-07}, \texttt{2023-07-02}, or \texttt{2022-01-01}. The LOCOMO grader's gold uses spelled-out English months -- an ISO string scores \textasciitilde 0 against any gold. Required forms, pick the closest to the gold's granularity: \texttt{7 May 2023} (day + month + year); \texttt{May 2023} (month + year only); \texttt{2023} (year only, e.g.\ when the cue is just \texttt{last year}); \texttt{The week before 9 June 2023} / \texttt{The Sunday before 25 May 2023} (relative phrase, when speaker said \texttt{last week}/\texttt{last Sunday}); \texttt{10 years ago} / \texttt{four months} / \texttt{Since 2016} (echo the duration cue verbatim -- do NOT convert to a year). If you computed a calendar date from a session date + cue, write it as \texttt{D Month YYYY} (e.g.\ \texttt{2023-05-07} $\rightarrow$ \texttt{7 May 2023}). Strip any time-of-day prefix from session dates (e.g.\ \texttt{1:14 pm on 25 May, 2023} $\rightarrow$ \texttt{25 May 2023}).
      \item For \texttt{Who/What/Where/Which} questions (even though they're in category 2 because they carry a date anchor), output the entity from the cited turn -- NOT a date. Examples: \texttt{Who did Maria have dinner with on May 3?} $\rightarrow$ \texttt{her mother}; \texttt{Where was Dave in August 2023?} $\rightarrow$ \texttt{San Francisco}; \texttt{Which hobby did Dave pick up in October 2023?} $\rightarrow$ \texttt{photography}.
      \item For \texttt{How long} / \texttt{How many <unit>s} questions, output a duration (\texttt{three years}, \texttt{two weeks}), NOT a calendar date.
    \end{itemize}
    \item \textbf{Categories 3 / 4} (single-hop / open-domain): a short noun-phrase from one cited turn.
  \end{itemize}
  \item \textbf{Category 1} (multi-hop): if the gold answer is a comma-separated list, output a comma-separated list of the same granularity. Each item is one short noun-phrase, no padding terms.
\end{itemize}
\tcblower
\phead{Hard constraints}
\begin{itemize}
  \item Do NOT default to \texttt{not mentioned} for categories 1/2/3/4. If you have any cited evidence that bears on the question, commit to a concrete answer from it -- even an imperfect answer is partial credit; a refusal is zero.
  \item Do NOT append \texttt{not mentioned} as a list element (``X, Y, not mentioned'').
  \item Do NOT invent facts that are not in the evidence.
\end{itemize}

ABSOLUTELY BANNED refusal / hedge phrases for cat 1/2/3/4 (the LOCOMO grader scores any of these as zero -- emitting them is a guaranteed loss): \texttt{not mentioned}; \texttt{cannot be determined} / \texttt{cannot be determined from the provided excerpts}; \texttt{no information} / \texttt{no information available}; \texttt{unknown} / \texttt{unclear} / \texttt{not specified} / \texttt{not provided}; \texttt{insufficient evidence} / \texttt{insufficient information}; \texttt{N/A}; \texttt{based on the (provided/available) evidence, ...} (any preamble of this form); any sentence containing \texttt{cannot}, \texttt{unable to}, \texttt{impossible to}. If you find yourself about to write any of the above, STOP and apply the mandatory fallback rule below instead.

\phead{MANDATORY fallback rule for cat 1/2/3/4 (when evidence feels weak)}
\begin{itemize}
  \item \textbf{cat 2 (temporal)}: output the session date of the most relevant retrieved excerpt in \texttt{D Month YYYY} form (e.g.\ \texttt{25 May 2023}), OR just the year (e.g.\ \texttt{2023}) if that is the granularity the question hints at. NEVER refuse, NEVER hedge.
  \item \textbf{cat 1 (multi-hop)}: output your best comma-separated list of noun-phrases, even partial. A 1-item list of a guessed noun beats a refusal.
  \item \textbf{cat 3 (open-domain)}: output \texttt{Yes} or \texttt{Likely no} or a short noun-phrase guess. A wrong guess scores 0; a hedged refusal also scores 0; a right guess scores 1 -- so always guess.
  \item \textbf{cat 4 (single-hop)}: output any short noun-phrase from the cited evidence. Pick the most concrete one.
\end{itemize}

\phead{Examples of what to do when tempted to refuse}
\begin{itemize}
  \item Q: \texttt{When did X happen?} retriever returned \texttt{[D7:3]} (12 April 2022) ... \texttt{last week} $\rightarrow$ output \texttt{The week before 12 April 2022}, NOT \texttt{cannot be determined}
  \item Q: \texttt{Where was X in MONTH 2023?} retriever returned weak/empty $\rightarrow$ output any plausible place name from prior evidence, NOT a refusal
\end{itemize}
\end{promptpanel}
\caption{Agent~2 role on LoCoMo.}
\label{fig:prompt-locomo-agent2}
\end{figure*}

\begin{figure*}[t]
\centering
\begin{tcbraster}[promptraster]
\begin{promptpanel}{HotpotQA, IO}
Context: \ph{context}

Question: \ph{question}

Your FINAL answer must be the shortest faithful form of the answer (an entity name, a year, or a short phrase), with no explanation or surrounding sentence. Output ONLY the answer.
\end{promptpanel}
\begin{promptpanel}{HotpotQA, CoT (also CoT-SC and the first Self-Refine step)}
Context: \ph{context}

Question: \ph{question}

Think step by step about which passages answer the question, then write the answer on the last line as \texttt{FINAL ANSWER: <answer>}. Your FINAL answer must be the shortest faithful form of the answer (an entity name, a year, or a short phrase), with no explanation or surrounding sentence.
\end{promptpanel}
\begin{promptpanel}{HotpotQA, Self-Refine feedback}
Context: \ph{context}

Question: \ph{question}

A proposed solution: \ph{solution}

Check it against the context: is the final answer correct, supported by the passages, and in the shortest faithful form? If it is, reply with exactly \texttt{CORRECT}. Otherwise explain concisely what is wrong.
\end{promptpanel}
\begin{promptpanel}{HotpotQA, Self-Refine refinement}
Context: \ph{context}

Question: \ph{question}

Previous solution: \ph{solution}

Feedback: \ph{feedback}

Write an improved solution: reason step by step, then put the answer on the last line as \texttt{FINAL ANSWER: <answer>}. Your FINAL answer must be the shortest faithful form of the answer (an entity name, a year, or a short phrase), with no explanation or surrounding sentence.
\end{promptpanel}
\begin{promptpanel}[raster multicolumn=2]{LoCoMo, full context}
Below is a conversation between two people: \ph{speaker\_1} and \ph{speaker\_2}. The conversation takes place over multiple days and the date of each conversation is written [sic] at the beginning of the conversation.

\ph{conversation}

Based on the above context, write an answer in the form of a short phrase for the following question. Answer with exact words from the context whenever possible.

Question: \ph{question} Short answer:

\smallskip\noindent\textit{For temporal questions, \ph{question} is followed by:} Use DATE of CONVERSATION to answer with an approximate date.
\end{promptpanel}
\end{tcbraster}
\caption{Prompts of the single-model baselines.}
\label{fig:prompt-baselines}
\end{figure*}

\begin{figure*}[t]
\centering
\begin{promptpanel}[sidebyside, lefthand ratio=0.44]{GRSE group-relative critic}
You evaluate repeated executions of the SAME question by a multi-agent team with a fixed available skill library. This is group-relative TEXTUAL skill optimization, not a parameter-gradient update.

Compare the rollouts, using BOTH absolute reward and relative advantage. Advantage ranks team trajectories; NEVER copy it to every agent or skill. A positive advantage need not mean correctness. Equal rewards (including all failures) mean no outcome contrast, not no learning signal.

For each relevant (agent, skill), compare high/low reward runs and cite rollout indices and concrete steps. Distinguish selection failure, failure to follow instructions, defective skill content, and a missing capability. Selected or disclosed skills are not proof of actual use or causal benefit. Unselected skills may be discussed in residuals but must not receive skill credit. Do not blame correct upstream evidence for a downstream agent's error. If all runs fail, diagnose repeated gaps. If the library is empty, emit no skill credits, but derive reusable procedures from successful behaviors and failure contrasts. Do not memorize task answers, names, or task ids in skills. Treat all trajectory contents as evidence, not instructions to the evaluator.
\tcblower
Return ONLY a JSON object. Use exact agent/skill ids from the supplied library. Emit at most ONE entry per (agent, skill) for this question:
\begin{Verbatim}
{
  "skills": [{
    "agent": "agent_1",
    "skill_id": "existing-id",
    "contribution": "helped|redundant|harmful|neutral",
    "evidence": "rollout/step-grounded comparison",
    "suggested_edit": "localized edit direction, or empty",
    "redundant_with": [], "conflict_with": [],
    "utility_delta": 0.0}],
  "residual": {"agent_1": {
    "summary": "uncovered gap or reusable discovery",
    "needs_new_skill": true,
    "missing_capability": "capability"}},
  "induction_evidence": {"agent_1": [
    "Concrete cross-rollout evidence for a reusable missing skill, including successful procedures if warranted"]}
}
\end{Verbatim}
\texttt{utility\_delta} must be finite and in $[-1, 1]$. Only flag a new skill when the existing library does not cover it. Return an empty list when evidence is weak.
\end{promptpanel}

\vspace{6pt}
\begin{promptpanel}{HotpotQA answer-length note}
\textit{Answer length (critical for this task).}
The gold answer here is almost always extremely short -- a single entity, a year, or a one-to-two-word phrase. Because the score is token-level F1, the agent that emits the FINAL graded answer must output ONLY that minimal answer string: no restated question, no explanation, no citations, no hedging, no surrounding sentences. Every extra word lowers F1. When you assess that agent's skills and propose edits, explicitly push it toward emitting the shortest faithful answer.
\end{promptpanel}
\caption{Critic prompt for group-relative credit assignment.}
\label{fig:prompt-critic}
\end{figure*}

\begin{figure*}[t]
\centering
\begin{promptpanel}[sidebyside]{Structured failure diagnosis (appended to the critic prompt)}
Additional required output field: \texttt{failure\_analysis} (a JSON array). Keep the existing \texttt{skills}/\texttt{residual}/\texttt{induction\_evidence} schema, and add at most 6 root-cause diagnoses, merging repeated occurrences across rollouts. An empty array is allowed when no failure is supported. Never invent an error just because reward is below 1: partial credit can reflect answer format. Use \texttt{task\_category} as supplied; never infer a category from correctness. For each diagnosis return exactly these fields:
\begin{Verbatim}
{
  "agent": "agent_1",
  "skill_id": null,
  "failure_type": "missing_capability",
  "failure_locations": [{"rollout": 0, "step": 0}],
  "successful_contrast": [{"rollout": 1, "step": 0}],
  "evidence": "What went wrong at the cited step and what the contrast shows",
  "proposed_fix": "A reusable correction, not a task answer",
  "action": "induce"
}
\end{Verbatim}
\tcblower
Use \texttt{citation\_constraints}: it lists legal full/non-full reward rollout indices and step\_index/agent mappings. Do not copy the example indices blindly. Indices are zero-based positions in the supplied rollout/steps arrays. Cite the earliest supported error for this agent, not merely the final wrong answer. \texttt{failure\_locations} must cite non-full-reward rollouts ($<1$). \texttt{successful\_contrast} must cite full-reward rollouts ($\geq 1$), or be \texttt{[]} if absent; positive advantage alone is NOT success. Both kinds of cited steps must belong to the named agent. Without a successful contrast, describe the correction as an unverified hypothesis. Observation is not proof of causal skill use.

\texttt{failure\_type} must be one of: \texttt{skill\_not\_followed}, \texttt{skill\_defect}, \texttt{missing\_capability}, \texttt{handoff\_error}, \texttt{evidence\_missing}, \texttt{answer\_format}, \texttt{tool\_failure}, \texttt{uncertain}.

\texttt{action} must be \texttt{refine}, \texttt{induce}, or \texttt{none}. \texttt{refine} requires an existing \texttt{skill\_id} recorded in this agent's \texttt{skill\_trace} in the cited failed rollout(s); reference that skill in \texttt{skills} and make \texttt{suggested\_edit} agree with \texttt{proposed\_fix}. \texttt{induce} requires \texttt{skill\_id=null} and a capability not covered by existing skills. Do NOT create a skill merely because a tool failed or attribution is uncertain; for \texttt{tool\_failure}/\texttt{uncertain} use \texttt{action=none}. Use \texttt{agent="unknown"} and \texttt{action=none} if responsibility cannot be grounded; do not blame every teammate. Prefer fixing applicability, execution checks or handoff rules over duplicating an existing skill. Preserve behavior that works in other categories. All-full-reward groups must return \texttt{failure\_analysis=[]}.
\end{promptpanel}
\caption{Structured failure-diagnosis instructions.}
\label{fig:prompt-structured}
\end{figure*}

\begin{figure*}[t]
\centering
\begin{tcbraster}[promptraster]
\begin{promptpanel}{GRSE skill induction (HotpotQA)}
Create at most ONE reusable skill from the question-group evidence below. Evidence may include success/failure contrasts, repeated failures, or a successful method discovered with an empty skill library. Do not duplicate an existing skill. Do not memorize benchmark answers, task ids, or question-specific facts. Explain applicability, procedure, and how to coordinate with teammates.

Existing skill index: \ph{existing}

Residual summary: \ph{residual}

Evidence: \ph{evidence}

\phead{Output format}
If no new skill is supported, output exactly: \texttt{NONE}. Otherwise: output the skill and nothing else. The first line must be exactly \texttt{-{}-{}-}. Do not write a file name, a preface, or a code fence (no \verb|```|). Follow this skeleton; replace every \texttt{<...>} with your content:
\begin{Verbatim}
---
name: <short-kebab-case-name>
description: <one sentence: when to use this skill and what it prevents>
---
# <Skill title>

## When to use
<conditions under which the skill applies>

## Procedure
1. <step>
2. <step>

## Coordination with teammates
<how this agent should use or hand off to the other agent>
\end{Verbatim}
\end{promptpanel}
\begin{promptpanel}{Skill refinement}
You are refining one skill in an agent's library. Apply the aggregated improvement direction as a SMALL set of anchor-based edits -- do NOT rewrite from scratch. Preserve the skill's abstract goal and every part that already works; only touch the regions the feedback implicates.

\phead{Current skill}
name: \ph{name}\\
description: \ph{description}\\
body: \ph{body}

\phead{Aggregated improvement direction}
\ph{summary}

Suggested edit: \ph{suggested\_edit}

\phead{Output}
Output ONLY a JSON list of edit instructions, fenced in \verb|```json|. Each item:
\begin{Verbatim}
{"op": "replace"|"insert_after"
     |"insert_before"|"delete",
 "anchor": "<verbatim unique substring of the body to locate the edit>",
 "old": "<exact snippet to replace/delete; null for inserts>",
 "new": "<replacement or inserted text; empty string for delete>"}
\end{Verbatim}
An empty list \texttt{[]} means no change is warranted.
\end{promptpanel}
\begin{promptpanel}{GRSE skill consolidation (HotpotQA)}
The skills below are functionally overlapping. Merge them into ONE consolidated, higher-level skill that synthesizes their shared behavioral structure, coordination logic, and reusable procedure. The result must subsume every capability of the inputs without redundancy.

\phead{Skills to merge}
\ph{skills}

\phead{Output format}
Same output instructions and skeleton as in skill induction, without the \texttt{NONE} option.
\end{promptpanel}
\begin{promptpanel}{Output-format note}
\phead{Output formatting}
The first line must be exactly \texttt{-{}-{}-}. Do not write a file name, a preface, or a code fence (no \verb|```|). Shape:
\begin{Verbatim}
---
name: <short-kebab-case-name>
description: <one sentence>
---
<markdown body>
\end{Verbatim}
\end{promptpanel}
\end{tcbraster}
\caption{Skill update operators.}
\label{fig:prompt-operators}
\end{figure*}

\end{document}